%% file: main.tex
\PassOptionsToPackage{table,xcdraw}{xcolor}
\documentclass[10pt,twocolumn,letterpaper]{article}

\usepackage[pagenumbers]{iccv} 

\usepackage{graphicx}
\usepackage{multirow}


\definecolor{iccvblue}{rgb}{0.21,0.49,0.74}
\usepackage[pagebackref,breaklinks,colorlinks,allcolors=iccvblue]{hyperref}

\usepackage{rotating}
\usepackage{array}
\usepackage{graphicx}
\usepackage{caption}
\usepackage{subcaption}
\usepackage{algorithm}
\usepackage{algpseudocode}

\def\paperID{5724} 
\def\confName{ICCV}
\def\confYear{2025}

\title{High-Resolution Image Synthesis via Next-Token Prediction}

\author{
    Dengsheng Chen$^{1}$ \quad
    Jie Hu$^{1,2}$ \quad
    Tiezhu Yue$^{1}$ \quad
    Xiaoming Wei$^{1}$ \quad
    Enhua Wu$^{2}$\thanks{This work is supported in part by NSFC Grants (62332015).} \\
    $^{1}$Meituan \quad
    $^{2}$Key Laboratory of System Software (Chinese Academy of Sciences) and \\
    State Key Laboratory of Computer Science, Institute of Software, Chinese Academy of Sciences \\
    {\tt\small \{chendengsheng, weixiaoming\}@meituan.com, \{hujie, weh\}@ios.ac.cn}
}

\begin{document}
\maketitle
\input{sec/abstract}
\input{sec/intro}

\input{sec/method}

\input{sec/exp}
\input{sec/con}

\input{sec/X_suppl}

{
    \small
    \bibliographystyle{ieeenat_fullname}
    \bibliography{main}
}

\end{document}

%% file: sec/abstract.tex
\begin{abstract}
Recently, autoregressive models have demonstrated remarkable performance in class-conditional image generation. However, the application of next-token prediction to high-resolution text-to-image generation remains largely unexplored. In this paper, we introduce \textbf{D-JEPA$\cdot$T2I}, an autoregressive model based on continuous tokens that incorporates innovations in both architecture and training strategy to generate high-quality, photorealistic images at arbitrary resolutions, up to 4K.
Architecturally, we adopt the denoising joint embedding predictive architecture (D-JEPA) while leveraging a multimodal visual transformer to effectively integrate textual and visual features. Additionally, we introduce flow matching loss alongside the proposed Visual Rotary Positional Embedding (VoPE) to enable continuous resolution learning.
In terms of training strategy, we propose a data feedback mechanism that dynamically adjusts the sampling procedure based on statistical analysis and an online learning critic model. This encourages the model to move beyond its comfort zone, reducing redundant training on well-mastered scenarios and compelling it to address more challenging cases with suboptimal generation quality.
For the first time, we achieve state-of-the-art high-resolution image synthesis via next-token prediction.
\end{abstract}

%% file: sec/intro.tex
\section{Introduction}

\input{figure/new_teaser}

In recent years, diffusion models have become the dominant approach for generating high-resolution images and videos from natural language inputs, demonstrating exceptional generalization capabilities~\citep{saharia2022photorealistic, ramesh2022hierarchical, podell2023sdxl, dai2023emu, esser2023structure, blattmann2023align, betker2023improving, blattmann2023stable, singer2022makeavideo, esser2024scaling, song2019generative, song2020score, ho2020denoising, dhariwal2021diffusion, rombach2022high, song2020denoising, lipman2022flow, liu2022flow, karras2022elucidating, gao2024lumina, zhuo2024lumina}. At the same time, the rise of autoregressive large language models~\citep{radford2018improving, radford2019language, brown2020language, ouyang2022training, achiam2023gpt, anil2023palm, chowdhery2023palm, hoffmann2022training, touvron2023llama, touvron2023llama2, le2023bloom, sun2021ernie, bai2023qwen, yang2024qwen2, team2023gemini, raffel2020exploring, yang2019xlnet} has marked a new era in artificial intelligence, leading to significant advances in artificial general intelligence~(AGI) due to their unparalleled versatility and generality.

The success of language models has similarly catalyzed advancements in image generation~\citep{ramesh2021zero, yu2021vector, yu2022scaling, sun2024autoregressive, liu2024lumina}. Recent works, such as D-JEPA~\citep{chen2024denoising}, MAR~\citep{li2024autoregressive}, and VAR~\citep{tian2024visual}, suggest that autoregressive models can achieve generative performance that rivals or even exceeds that of diffusion models in class-conditioned image synthesis on ImageNet~\citep{russakovsky2015imagenet}. However, despite their advantages in prompt adherence and computational efficiency over diffusion models~\citep{kilian2024computational}, autoregressive models still face challenges in generating high-resolution images with fine-grained textures and overall visual fidelity.

To further advance autoregressive models for image generation and to foster the development of unified multi-modal models, we focus on two crucial aspects in this work: model architecture and training strategy.

From an \textbf{architectural perspective}~(Sec. \ref{sec: model architecture}), we adopt the denoising joint embedding predictive architecture (D-JEPA)~\citep{chen2024denoising}, building on its success in integrating representational learning to enhance model performance~\citep{yu2024representation}. We also draw inspiration from the successful design principles of diffusion models, incorporating a multimodal visual transformer block, initially introduced by~\citeauthor{esser2024scaling}, to ensure effective fusion of visual and textual features. Additionally, we leverage the more robust flow matching loss~\citep{liu2022flow} to restore tokens into image patches. 
Furthermore, we identify that the conventional patchification operation, which segments continuous images into discrete blocks using RoPE~\citep{su2024roformer}, limits the model's ability to handle continuous resolutions and varying aspect ratios. To address these challenges, we introduce the visual rotary positional embedding (VoPE), enabling the model to learn across continuous resolutions and dynamic aspect ratios.

In terms of \textbf{training strategy}~(Sec. \ref{sec: training strategy}), we propose a novel data feedback mechanism to optimize resource efficiency. Traditional data curation typically involves preprocessing steps, such as filtering low-quality images~\citep{he2022rethinking} and refining prompts using multimodal models~\citep{li2024hunyuan}. While these methods improve data quality, they often introduce biases and fail to adapt to evolving data distributions, particularly in large-scale datasets. Fine-tuning techniques, such as reinforcement learning from human feedback~(RLHF)~\citep{xu2024imagereward, liang2024rich} and direct preference optimization~(DPO)~\citep{rafailov2024direct, wallace2024diffusion}, offer post-hoc adjustments but exhibit inconsistent effectiveness.
In contrast, our data feedback mechanism dynamically adjusts the data distribution in real-time based on statistical analysis and model performance, as assessed by the critic model, during training. Specifically, the critic model is trained to evaluate the model's performance on a sampled data point, providing continuous feedback to guide training.

By integrating the data-feedback training strategy into D-JEPA$\cdot$T2I, we achieve state-of-the-art performance in high-resolution image synthesis via next-token prediction. Our approach has been validated on the T2I-CompBench~\citep{huang2023t2i}, GenEval~\citep{ghosh2024geneval}, and GenAI-Bench~\citep{li2024genai} benchmarks, as well as through human evaluations.

%% file: figure/new_teaser.tex
\begin{figure*}[ht]
    \centering
    \includegraphics[width=\linewidth]{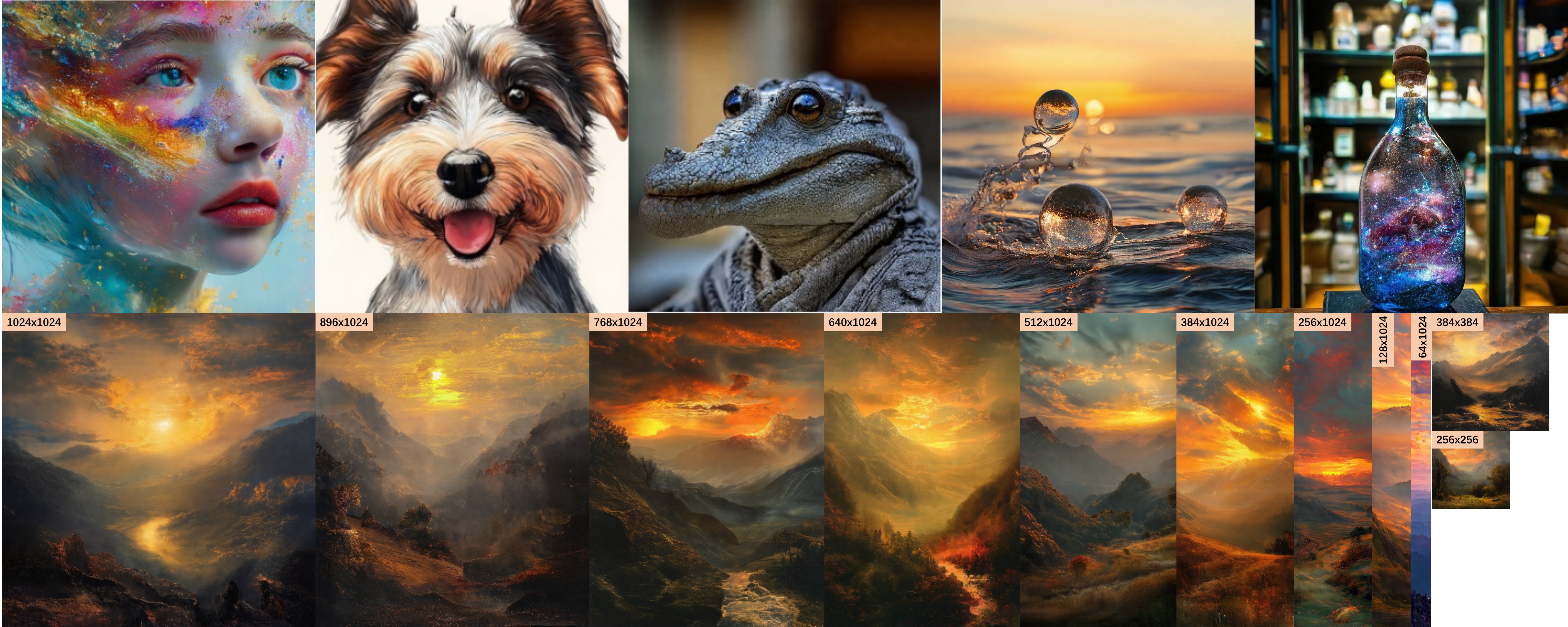}
    \caption{D-JEPA$\cdot$T2I can accurately generate high-fidelity, high-resolution images across various aspect ratios. Refer to the supplementary materials for 4K resolution samples and additional qualitative results.}
    \label{fig: new teaser}
    \vspace{-12pt}
\end{figure*}

%% file: sec/method.tex
\section{Model Architecture}  
\label{sec: model architecture}

The denoising joint embedding predictive architecture (D-JEPA)~\citep{chen2024denoising} builds upon the visual transformer~\citep{dosovitskiy2020image} and models the token distribution $p(x_i | z_i)$, where $z_i$ represents the predicted features of each token. It employs a combination of feature prediction loss $\mathcal{L}_{\text{pred}}$ and diffusion loss $\mathcal{L}_{\text{diff}}$. While D-JEPA has demonstrated strong performance in class-conditioned image generation, its applicability remains limited to fixed-resolution image synthesis, typically at $256 \times 256$ or $512 \times 512$.  

In this work, we introduce D-JEPA$\cdot$T2I, extending D-JEPA to high-resolution text-to-image generation, as shown in Fig.~\ref{fig: djepa t2i short}. To achieve this, we adapt a multimodal visual transformer, building upon~\citep{esser2024scaling}, to more effectively integrate textual and visual features (Sec.~\ref{sec: mvt}). Additionally, we replace the diffusion loss $\mathcal{L}_{\text{diff}}$ with a more flexible and faster-converging flow matching loss $\mathcal{L}_{\text{flow}}$ (Sec.~\ref{sec: fml}). Finally, we propose VoPE, a novel positional embedding for continuous-resolution learning (Sec.~\ref{sec: vope}).

\subsection{Multimodal Visual Transformer}
\label{sec: mvt}
The multimodal visual transformer draws inspiration from the design of the multimodal diffusion backbone, initially proposed by \citet{esser2024scaling}. The core idea is that text and image embeddings are conceptually quite different, necessitating the use of two separate sets of weights for the two modalities. This approach is equivalent to having two independent transformers for each modality while concatenating their sequences for the attention operation. This setup allows both representations to operate within their own spaces while still incorporating information from the other.\footnote{Refer to the supplementary materials for an illustration of the architecture.} \citet{dehghani2023scaling} observe that the training of large vision transformer models diverges because the attention entropy grows uncontrollably. To avoid this, they propose normalizing Q and K before the attention operation. We follow this approach and use RMSNorm~\citep{zhang2019root} with a learnable scale in both streams of D-JEPA$\cdot$T2I architecture. The additional normalization prevents attention logit growth instability, confirming findings by previous works\citep{dehghani2023scaling, wortsman2023smallscale, esser2024scaling}, and enables efficient training at bf16-mixed precision~\citep{chen2019bfloat16} when combined with the AdamW~\citep{loshchilov2017fixing} optimizer.

The \textit{primary distinction between the multimodal visual transformer and the multimodal diffusion backbone} is that the former does not require handling the additional timestep $t$ introduced by the diffusion process. This omission sidesteps adaptive layer norm~\citep{perez2018film}, which is essential for diffusion models built on top of DiT~\citep{peebles2023scalable}. 

For textual tokens, we follow the design principles of large language models and use RoPE~\citep{su2024roformer} as positional encoding. For visual tokens, we introduce VoPE, a positional encoding specifically designed for visual features, which we detail in Sec. ~\ref{sec: vope}.
\input{figure/djepa_t2i_short}

\subsection{Flow Matching Loss}
\label{sec: fml}

Flow matching~\citep{lipman2022flow, albergo2023stochastic,ma2024sit, mo2024freecontrol} emerges as a simple alternative that linearly interpolates between noise and data along a straight line. In this context, we adhere to the flow matching formulation presented in \citet{gao2024lumina} for modeling the token distribution $p(x_i|z_i)$. More specifically, given the data \( x_i \sim p(x_i|z_i) \) and Gaussian noise \( \epsilon \sim \mathcal{N}(0, I) \), we define an interpolation-based forward process:
$$
x_i^t = \alpha_t x_i + \beta_t \epsilon,
$$
where \( \alpha_0 = 0 \), \( \beta_t = 1 \), \( \alpha_1 = 1 \), and \( \beta_1 = 0 \). This interpolation for \( t \in [0,1] \) bridges \( x_i^0 = \epsilon \) and \( x_i^1 = x_i \). Similar to the diffusion schedule, this interpolation schedule offers flexible choices of \( \alpha_t \) and \( \beta_t \). In our framework, we adopt a linear interpolation schedule between noise and data for its simplicity: $x_i^t = t x_i + (1-t) \epsilon$. This formulation represents a uniform transformation with constant velocity between the data and noise. The corresponding time-dependent velocity field is defined as:
$$
v_t(x_i^t, z_i) = \dot{\alpha}_t x_i + \dot{\beta}_t \epsilon = x_i - \epsilon,
$$
where \( \dot{\alpha} \) and \( \dot{\beta} \) denote the time derivatives of \( \alpha \) and \( \beta \). This time-dependent velocity field \( v : [0,1] \times \mathbb{R}^d \to \mathbb{R}^d \) defines an ordinary differential equation known as the Flow ODE:
$$
dx_i = v_t(x_i^t, z_i) dt.
$$

We use \( \psi_t(x_i, z_i) \) to represent the solution of the Flow ODE with the initial condition \( \psi_0(x_i, z_i) = x_i \). 
By solving this Flow ODE from \( t = 0 \) to \( t = 1 \), we transform noise into data samples using the approximated velocity fields \( v_{\theta}(x_i^t, t, z_i) \). Similar to the approaches of \citet{chen2024denoising} and \citet{li2024autoregressive}, $v_{\theta}$ is implemented with a small denoising MLP~\citep{li2024autoregressive}.

During training, the flow matching objective directly regresses to the target velocity for each token:
\begin{equation}
\mathcal{L}_{\text{flow}}(x_i, z_i) = \int_0^1 \mathbb{E}\left[ \parallel v_{\theta}(x_i^t, t, z_i) - (x_i - \epsilon) \parallel^2 \right] dt,
\label{eq: flow matching loss}    
\end{equation}
which is termed the conditional flow matching loss, sharing similarities with the noise prediction or score prediction losses in diffusion models~\citep{lipman2022flow}.

\subsection{VoPE for Continuous Resolution Learning}
\label{sec: vope}

\input{figure/vope}
Empirically, autoregressive models face challenges when generating images with arbitrary resolutions and aspect ratios, mainly due to the absence of appropriate visual positional embeddings. Both sinusoidal positional encoding and rotary positional embedding have limitations: the former cannot ensure positional consistency when images are cropped or padded, and neither can maintain positional information consistency across different scales of the same image. Consequently, operations such as cropping, padding, or scaling an image can lead to models receiving completely different positional information if existing positional embedding schemes are applied directly. 

These limitations significantly impact the learning of token distributions, potentially leading to flaws in generated high-resolution images. Here, we propose VoPE, a novel positional embedding for continuous resolution learning.

\paragraph{Visual-rotary Positional Embedding.}
\input{figure/data_feedback}
Visual-rotary positional embedding (VoPE) is inspired by biological vision and camera imaging principles. VoPE assumes that all objects are projected within a field of view with a fixed resolution of $g \times g$. For an image with resolution $W \times H$, any pixel at coordinate $(w, h)$ can be normalized as:
\[
\left\{
\begin{matrix}
(\frac{1}{\rho}(w+b), \frac{1}{\rho}(h)), & W \le H \\
(\frac{1}{\rho}(w), \frac{1}{\rho}(h+b)), & W > H
\end{matrix}
\right.
\]
where $\rho = \max(W, H) / g$ is the resolution density, and \( b = \mathrm{abs}(W - H) / 2 \) is the relative positional offset. This normalization centers images on the grid \(g \times g\), facilitating efficient learning from images of varying resolutions.

Consider a single row of tokens where \( q_m \) and \( k_n \) are tokens from query and key sequences, respectively. \citet{su2024roformer} aim to implement relative positional encoding through absolute positional encodings with RoPE. VoPE targets the dot product between \( q_m \) and \( k_n \) to incorporate the relative positional information \(\frac{1}{\rho}(m-n)\) while maintaining translation invariance. Thus, the inner product encodes positional information in a relative \textit{pixel normalized} manner:
\[
\langle f_q(x_m, \frac{1}{\rho}(m+b)), f_k(x_n, \frac{1}{\rho}(n + b)) \rangle = g(x_m, x_n, \frac{1}{\rho}(m-n)).
\]

The aim is to determine an equivalent encoding mechanism for \( f_q(x_m, \frac{1}{\rho}(m+b)) \) and \( f_k(x_n, \frac{1}{\rho}(n+b)) \) that satisfies the above relation. Referring to \citet{su2024roformer}, the functions \( f \) and \( g \) that meet the relationship can be defined as follows when the feature dimension \( d = 2 \):
\begin{align*}
f_q(x_m, \frac{1}{\rho}(m + b)) &= (\mathbf{W}_q x_m)e^{i [\frac{1}{\rho}(m + b)] \theta} \\
f_k(x_n, \frac{1}{\rho}(n + b)) &= (\mathbf{W}_k x_n)e^{i [\frac{1}{\rho}(n+b)] \theta} \\
g(x_m, x_n, \frac{1}{\rho}(m-n)) &= \text{Re}[(\mathbf{W}_q x_m)(\mathbf{W}_k x_n)^{\star}e^{i[\frac{1}{\rho}(m-n)]\theta}] ,
\end{align*}
where \( \text{Re}[\cdot] \) denotes the real part, and \( (\mathbf{W}_k x_n)^\star \) is the complex conjugate of \( (\mathbf{W}_k x_n) \). The constant \( \theta \in \mathbb{R} \) is preset non-zero. When \( d > 2 \), \( \theta_j = \omega^{-\frac{2(j-1)}{d}}, j \in [1, 2, \dots, d/2] \), where \( \omega \) is a preset base frequency. For larger \( d \), the derivations and expressions of \( f_q(x_m, \frac{1}{\rho}(m + b)) \) and \( f_k(x_n, \frac{1}{\rho}(n + b)) \) are consistent with RoPE~\citep{su2024roformer} and are not reiterated here.

\paragraph{Comparison between VoPE and RoPE.}
During sampling, RoPE requires adjusting the base frequency $\omega$ for higher resolution images. For instance, with a 256 token training length and a target of 512 tokens, the NTK-Aware Scaled RoPE approach~\citep{peng2023ntk} is used, where $\omega^{\prime} = \frac{\omega}{2}$. Although effective for long texts, this causes discrepancies in positional information between training and sampling phases, as shown in Fig.~\ref{fig: rope decay}. These discrepancies are detrimental for image generation, which is sensitive to token boundary information. \citet{gao2024lumina} noted that this method leads to blurry, repetitive images in higher resolution image generation, akin to issues in positional interpolation or extrapolation. Thus, RoPE is unsuitable for arbitrary resolution image generation, especially at higher resolutions.
In contrast, VoPE ensures that pixel normalization maintains consistent positional information across resolutions during training and sampling. This is achieved by normalizing images to a $g \times g$ grid via \(\rho\), without changing the base frequency \(\omega\), regardless of resolution. As shown in Fig.~\ref{fig: vope decay}, images at different resolutions with VoPE use the same relative positional curve at varying resolution densities \(\rho\). Notably, at \(\rho = 1/4\), the curve closely matches that of \(\rho = 1.0\), implying that to generate $4096 \times 4096$ images, training at $1024 \times 1024$ with \(\rho = 1/4\) suffices.\footnote{This assumes a VAE encoding stride of 8 and a patch size of 2.}
Beyond enabling different resolution image generation via \(\rho\), VoPE allows layout adjustments in generated images with varying aspect ratios by adjusting the relative position bias \(b\). \textit{Refer to the supplementary materials for further results.}

\subsection{High-resolution Image Sampling}
\label{sec: inference and sampling}

For evaluating generative models in generalized next-token prediction, we employ an iterative sampling strategy similar to those used in \citet{chang2022maskgit, li2024autoregressive}, as outlined in Algo.~\ref{alg: sampling}. This strategy gradually decreases the masking ratio from \(1.0\) to \(0.0\) following a cosine schedule, typically using \(64\) autoregressive steps for sampling an image with a resolution of $256\times256$. Empirically, we find that even for higher-resolution images (such as 2K or 4K), satisfactory sampling quality can be achieved within approximately 100 autoregressive steps. D-JEPA$\cdot$T2I follows the approach of \citet{chen2024denoising}, utilizing fully randomized orderings to determine the next set of tokens to predict. This design effectively enhances the diversity of the generated samples.

\input{algo/sampling}

\section{Training Strategy}
\label{sec: training strategy}

While \citet{esser2024scaling} and \citet{li2024hunyuan} meticulously curate high-quality training data, they lack an in-depth examination of the actual sampled training data. To address this, we propose a statistical analysis and critic model sampling approach for more refined data selection. We refer to this novel training strategy as \textbf{data feedback}, and the training pipeline is depicted in Fig.~\ref{fig: data-feedback}. 

\subsection{Statistical Analysis Sampling.}
It is worth noting that the curated training data does not always match the sampled training data and does not guarantee an optimal text-to-image model due to inherent data biases in natural distributions and the potential loss of long-tail data due to sampling. \textbf{(a) Data bias in natural distribution.} The distribution of training data obtained through curation rules typically presents significant biases. Since a large volume of training data is often sourced from publicly available internet data, the overall distribution, even post-curation, shows strong concentration regarding resolution and semantic notions.~\footnote{Please } This can result in redundant data, thereby reducing the model's overall performance. \textbf{(b) Under-sampling of long-tailed data.} Directly training with vast datasets (usually ranging from tens of millions to billions of image-text pairs) can potentially lead to the loss of long-tail data. In small-scale training, such as with ImageNet, the entire dataset is extensively traversed (e.g., thousands of iterations in \citet{peebles2023scalable}), providing the model ample opportunity to learn the distribution of all training data. However, when the dataset scales up to billions, each data point might only be traversed a few times, and often the entire dataset might not be fully traversed even once. Resuming training typically recovers prior model parameters and optimizer states but struggles to track previously trained data indices, exacerbating the loss of long-tail data.

\paragraph{Statistical Analysis Sampling.} 
\label{para: sas}
To address these challenges, we perform statistical analysis on each batch of sampled data, examining features such as resolution, prompt composition, and style tags, and feed these features back into the sampling process for subsequent iterations in real-time. When sampling data, we introduce two distinct strategies to handle transformable and non-transformable attributes. \textit{For transformable attributes} such as image resolution and aspect ratio, which can be adjusted without quality degradation, we utilize a truncated normal distribution sampling strategy, denoted as $\mathrm{trunc\_norm}(\mu,\sigma,a,b)$. We initially sample the desired data parameters, such as the target resolution. If the current data can be transformed to align with these specified parameters, we proceed with the transformed data; otherwise, we discard this sample and attempt resampling. In contrast, \textit{for non-transformable attributes}, we decide whether to use the current data based on a predetermined sampling frequency specific to each attribute. If the data exceeds this frequency, there is a 50\% probability it will be discarded. Although this approach may not strictly adhere to the exact predetermined sampling frequencies, it significantly enhances data utilization, particularly beneficial when dealing with the sampling of long-tail distributions.

Based on statistical analysis sampling, we can achieve a more reliable and balanced selection, ensuring uniform sampling of all data types as expected.  

\subsection{Critic Model Sampling}

Does uniformly sampling all types of data necessarily lead to a well-trained generative model? The answer is clearly no. While diverse data distribution ensures a broad representation within the dataset, diversity alone is merely a fundamental prerequisite for training a model with general capabilities. Critic model sampling is designed to address the inherent weaknesses of the model during the training process.

In the early training stages, we employ statistical analysis sampling to expose the model to a wide variety of data, helping it establish a foundational knowledge system. However, in practice, we observe that the model quickly learns to generate certain types of content while struggling with others. Some types of data may take significantly longer to learn, or in some cases, the model may never master them. Critic model sampling aims to identify the types of content that the model finds challenging and prioritize them during training. By doing so, we continuously refine the model's abilities and mitigate its weaknesses.

\input{figure/critic_model}

A key aspect of this approach involves the online learning of the critic model. As illustrated in Fig.~\ref{fig: critic model}, we follow a structured pipeline for training the critic model. \textbf{(a) Data preparation.} First, we collect a sufficiently large set of prompts that cover a comprehensive range of scenarios (typically in the thousands). Next, we load the current T2I model weights and generate a substantial number of synthetic images for each prompt to form a reference set. Additionally, we retrieve several images from the training dataset based on text similarity with each prompt to construct the training set. \textbf{(b) Data annotation.} \textit{For the reference set}, in the early training stages—when the model's performance is still suboptimal—we utilize automated evaluation metrics such as T2I CompBench~\citep{huang2023t2i} to assess whether generated images meet the fundamental requirements of the prompts. Images that meet the criteria are labeled as positive samples, while those that fail are labeled as negative samples. As training progresses and the model's output quality improves, we introduce human evaluation. At this stage, experts assess each generated image from a professional photography perspective, considering aspects such as logical consistency and realism. Images that meet the quality criteria are labeled as positive samples, while those that do not are marked as negative. \textit{For the training set}, after completing the reference set annotations, we compute the feature similarity between images in the training set and their corresponding reference set images. Images with high feature similarity inherit the reference set labels, while those with low similarity are discarded. Since the model is still relatively weak in the early stages, we adopt a lower similarity threshold for automatically labeled reference images. In later stages, when human annotations are available, we raise the threshold to ensure a sufficiently large and high-quality training dataset. \textbf{(c) Training the critic model.} We employ a ResNet-18 model~\citep{he2016deep}, modifying it into a binary classification network. For the first round of critic model training, we initialize it with ImageNet pre-trained weights collected by \citeauthor{rw2019timm}. In subsequent updates, we use the previous version of the critic model as the initialization. Since the T2I model continuously evolves, the critic model must also be updated accordingly. In our implementation, we introduce critic model sampling after the first 100K iterations of the T2I model. Thereafter, we update the critic model every 40K iterations. The training process follows standard ImageNet classification strategies and data augmentation techniques~\citep{rw2019timm}. Given that the dataset consists of tens of thousands of samples, the computational overhead of training the critic model is negligible.

\paragraph{Critic Model Sampling.}
\label{para: cms}
Once training is complete, we validate the samples obtained from statistical analysis sampling. The critic model assigns each sample a probability of being rejected (\textit{i.e.}, the likelihood that the model struggles with this type of content). Since we aim to sample more challenging cases where the model performs poorly, a higher rejection probability indicates that the model handles the sample well and should be discarded. Conversely, lower-probability samples are prioritized for training.

\paragraph{Comparison between critic model sampling and fine-tuning techniques.}  
Fine-tuning techniques, such as reinforcement learning from human feedback (RLHF)~\citep{xu2024imagereward, liang2024rich} and direct preference optimization (DPO)~\citep{rafailov2024direct, wallace2024diffusion}, provide post-hoc adjustments but often exhibit inconsistent effectiveness. Our proposed critic model sampling can be seen as an online counterpart to fine-tuning techniques.  
Moreover, the utilization of human annotation data differs between these approaches. In the fine-tuning stage, human annotators typically collect a new source of high-quality training data to directly improve the T2I model. In contrast, critic model sampling requires only a few thousand synthetic images to be annotated with positive and negative labels, which are then used solely for training the critic model—without necessitating additional new training data sources.  In summary, critic model sampling focuses on improving the utilization of existing data, whereas fine-tuning techniques emphasize incorporating new data sources to enhance model performance.

%% file: figure/djepa_t2i_short.tex
\begin{figure}
    \centering
    \includegraphics[width=0.8\linewidth]{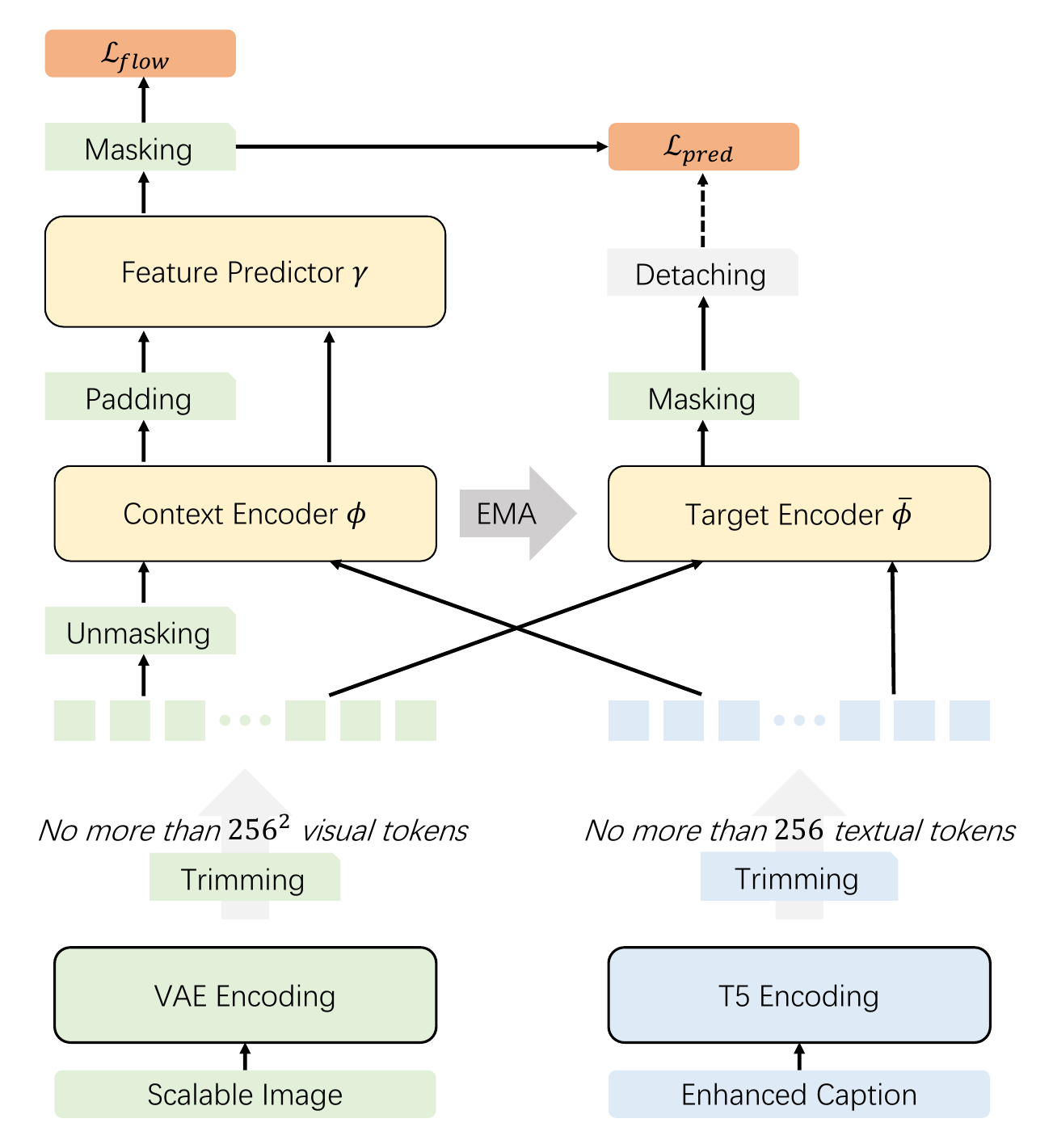}
    \caption{Denoising with a Joint-Embedding Predictive Architecture for text-to-image synthesis. We employ T5-XXL~\citep{chung2024scaling} as the text encoder, and the KL-VAE pretrained by \citet{esser2024scaling} as the image encoder. Both textual and visual tokens are trimmed to no more than 256 and $256^2$ tokens for efficient training, respectively. The feature predictor $\gamma$, the context encoder $\phi$, and the target encoder $\bar{\phi}$ share the same network architecture, each consisting of several multimodal visual transformer blocks. The gradient is detached from the output of the target encoder $\bar{\phi}$, ensuring that it is only updated via exponential moving average~(EMA). Both the prediction loss $\mathcal{L}_{\text{pred}}$ and the flow matching loss $\mathcal{L}_{\text{flow}}$ are computed only for the masked visual tokens, following \citet{chen2024denoising}.}
    \label{fig: djepa t2i short}
\end{figure}

%% file: figure/vope.tex
\begin{figure}
    \centering
    \begin{subfigure}[b]{0.48\linewidth}
        \centering
        \includegraphics[width=\linewidth]{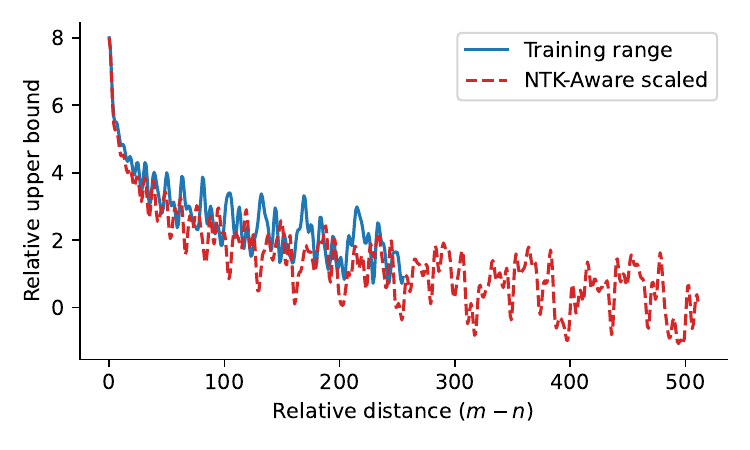}
        \caption{RoPE.}
        \label{fig: rope decay}
    \end{subfigure}
    \hfill
    \begin{subfigure}[b]{0.48\linewidth}
        \centering
        \includegraphics[width=\linewidth]{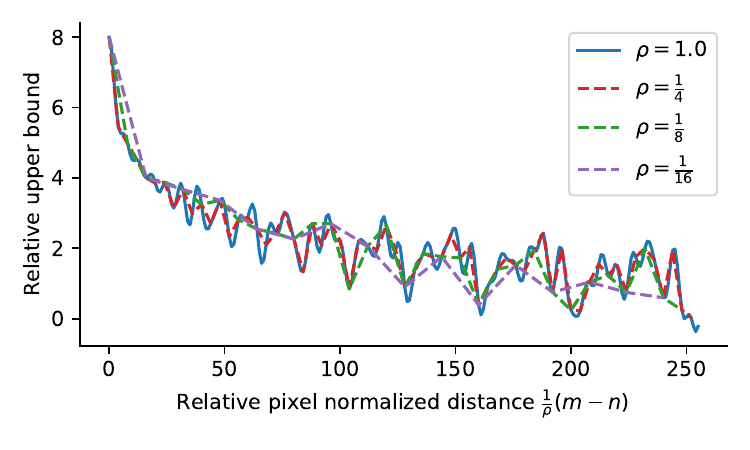}
        \caption{VoPE.}
        \label{fig: vope decay}
    \end{subfigure}
    \caption{Comparison of decay curves between RoPE and VoPE.}
    \label{fig: vope}
    \vspace{-12pt}
\end{figure}

%% file: figure/data_feedback.tex
\begin{figure*}[]
    \centering
    \includegraphics[width=\linewidth]{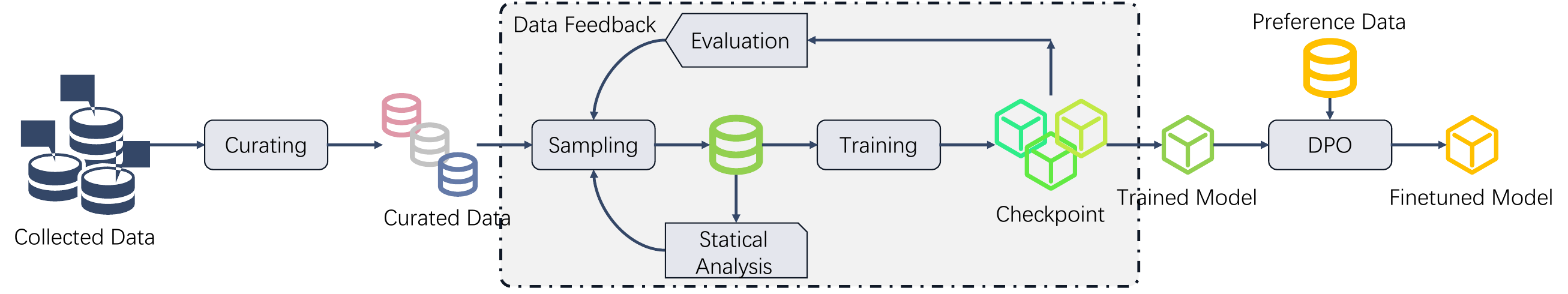}
    \caption{Training procedure incorporating data feedback. The evaluation result will be used to train the critic model.}
    \label{fig: data-feedback}
\end{figure*}

%% file: algo/sampling.tex
\begin{algorithm}[H]
  \begin{algorithmic}[1]
    \Require $T$: Number of auto-regressive steps, $N$: Total tokens to sample.
    \State \textbf{Initialize:} $\mathbb{X} \gets \emptyset$
    \For{$n$ in $\text{cosine-step-function}(T, N)$}
      \State $\mathbb{C} \gets \phi(\mathbb{X})$ \Comment{Encode the sampled tokens}
      \State $\mathbb{Z} \gets \gamma(\mathbb{C})$ \Comment{Predict features of unsampled tokens}
      \State $\{z_0, \ldots, z_n\} \sim \mathbb{Z}$ \Comment{Randomly select $n$ tokens}
      \State $\{x_0, \ldots, x_n\} \gets \text{denoise}(\epsilon_\theta, \{z_0, \ldots, z_n\})$
      \State $\mathbb{X} \gets \mathbb{X} \cup \{x_0, \ldots, x_n\}$ \Comment{Add the denoised tokens}
    \EndFor
    \State \textbf{Return:} $\mathbb{X}$
  \end{algorithmic}
  \caption{High-resolution image synthesis with D-JEPA$\cdot$T2I in generalized next-token prediction.}
  \label{alg: sampling}
\end{algorithm}

%% file: figure/critic_model.tex
\begin{figure}[]
    \centering
    \includegraphics[width=\linewidth]{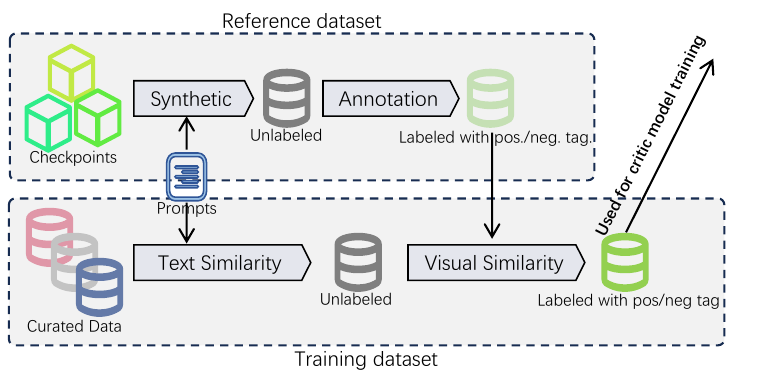}
    \caption{The pipeline to prepare training set for critic model.}
    \label{fig: critic model}
    \vspace{-12pt}
\end{figure}

%% file: sec/exp.tex
\section{Experiments}
\input{table/genai}

In this section, we succinctly outline the experimental configuration and training process of D-JEPA$\cdot$T2I. We evaluate the model performance through automated metrics, qualitative results, and human ratings. \textit{\textbf{Detailed experimental setups and crucial ablation studies} (concerning VoPE, data feedback, and more) are provided in the supplementary material.}

\subsection{Experiment Setup}

\paragraph{Dataset.} We employ an internally curated dataset comprising over 1 billion image-text pairs for training. Each image has a minimum shorter side length of 512 pixels. To maintain high quality, images with aesthetic scores below $5.0$ are excluded using the LAION-AI aesthetic predictor\footnote{\url{github.com/LAION-AI/aesthetic-predictor}}. Additionally, OCR tools filter out images containing text, which constrains the model from generating text but enhances its learning of other real-world concepts. English captions are generated for each image using InternVL2~\citep{chen2023internvl, chen2024far}, and enhanced with image tags (e.g., style tags, data source). Synthetic datasets, like JourneyDB~\citep{sun2024journeydb}, are carefully incorporated despite their efficacy in accelerating convergence~\citep{chen2023pixart, gao2024lumina}. We hypothesize that over-dependence on synthetic data might limit the model's output diversity, reducing data utility. Thus, synthetic data constitutes only about 5\% of the dataset, ensuring D-JEPA$\cdot$T2I generates authentic content, albeit with slower convergence.

\input{figure/human_preference}
\input{table/geneval_compbench_short}
\paragraph{Training.} The D-JEPA$\cdot$T2I model extends D-JEPA-H~\citep{chen2024denoising} and comprises 2.6 billion parameters. Training is strategically divided into two phases. \textit{Initially}, the model is trained on images no larger than $256 \times 256$ pixels\footnote{For images with resolutions greater than $256 \times 256$, we apply aspect ratio-preserving resizing.}, using a batch size of 2048 for 100k steps, with only statistical analysis sampling applied. This phase is crucial for enhancing textual concept comprehension. A cosine annealing schedule is employed, adjusting the learning rate from $1 \times 10^{-5}$ to $1 \times 10^{-6}$, with a warm-up of 10k steps. \textit{In the second phase}, the model trains on diverse image scales and resolutions, with batch sizes dynamically adjusted based on sampled resolutions from a truncated normal distribution (see Para.~\ref{para: sas}). Image resolutions progressively increase from $128$ to $1024$ pixels over the subsequent 500k iterations. Refer to the \textit{supplementary material} for more details on resolution sampling. During this phase, critic model sampling is introduced and updated every 40k iterations.  

Empirically, we find that with statistical analysis sampling alone, the model achieves an overall score of $0.50$ on GenEval within 100k training steps, matching the performance of SD2.0~\citep{rombach2022high} and PixArt-$\alpha$~\citep{chen2023pixart}. After introducing the critic model, trained with the automated evaluation-based trick, the model approaches its final performance at 200k steps, showing gradual improvement between 200k and 300k steps, ultimately reaching $0.66$—surpassing models of comparable scale. In the final 300k training steps, the reference set is labeled via human evaluation. While this stage does not significantly improve performance on standard benchmarks like GenEval~\citep{ghosh2024geneval}, it leads to substantial advancements in image realism and photographic style, as reflected in human ratings, which increased from a win rate of less than 20\% against Midjourney v6 to over 30\%.  

The full training process was conducted on 128 H800 GPUs over two weeks. The reference set required for updating the critic model was constructed using two different approaches based on the annotation method. During the automated annotation phase (100k–300k steps), we generated approximately 100k images at $256 \times 256$ resolution, as this stage primarily targets fundamental model capabilities. Since automated annotation mainly detects the presence of corresponding concepts, generating low-resolution images not only reduces inference time but also ensures comprehensive model evaluation. In the later stage (300k–600k steps), we generated around 10k images at resolutions ranging from $512$ to $1024$ pixels. At this point, the model had already acquired strong generative abilities, allowing human annotators to conduct meticulous evaluations and enforce stricter selection criteria, further driving model optimization.

Before conducting the final evaluation on human rating, we performed direct preference optimization~(DPO) training. This training utilized an additional ultra-high-quality dataset and was conducted for 20k steps. The primary objective of this stage was to refine the model's image style and texture, making them more closely resemble real photographic aesthetics. Notably, we did not attempt to enhance other aspects of the model, such as text generation quality.

\paragraph{Inference.} The D-JEPA$\cdot$T2I model can generate images at arbitrary resolutions and aspect ratios. For $256 \times 256$ resolution images used in quantitative evaluation, we set autoregressive steps to $T=64$. For higher resolutions, autoregressive steps are empirically tuned. Classifier-free guidance~\cite{ho2021classifier} enhances image quality, with hyperparameters optimized per benchmark. Denosing MLP diffusion steps are consistently set to $250$ across tasks.

\subsection{Automated Metric Evaluation}

We evaluate our method using automated metrics on prominent text-to-image benchmarks, including GenEval~\citep{ghosh2024geneval}, T2I-CompBench~\citep{huang2023t2i}, and GenAI-Bench~\citep{li2024genai}, which assess the model's capacity to generate prompt-reflective images. Table~\ref{tab: geneval compbench short} compares D-JEPA$\cdot$T2I against state-of-the-art diffusion and autoregressive models, both open-source and closed-source. The number of model parameters significantly influences performance; thus, comparisons among similar-scale models offer practical insights.

Per GenEval overall scores in Tab.~\ref{tab: geneval compbench short}, D-JEPA$\cdot$T2I surpasses other models within small and mainstream sizes. Compared with larger models, D-JEPA$\cdot$T2I (2.6B parameters) excels over Emu3~\citep{wang2024emu3} (8.0B) and Transfusion (7.3B), and is competitive with Fluid (10.5B). It also rivals commercial models like DALL$\cdot$E 3~\citep{betker2023improving} and Midjourney v6~\citep{midjourneyv6}.

Regarding T2I-CompBench++, D-JEPA$\cdot$T2I outperforms previous open-source works like PixArt-$\alpha$~\citep{chen2023pixart} and SDXL~\citep{podell2023sdxl}, achieving a pioneering level in the field. In Tab.~\ref{tab: genai}, D-JEPA$\cdot$T2I surpasses SD 3.0~\citep{esser2024scaling} in advanced prompt generation, maintaining a leading position. These findings robustly demonstrate that as an autoregressive model, D-JEPA$\cdot$T2I has achieved state-of-the-art performance in text-to-image tasks for the first time, exhibiting substantial potential.

\subsection{Qualitative Results}

Fig.~\ref{fig: new teaser} illustrates the versatile capabilities of images generated by D-JEPA$\cdot$T2I, which supports flexible resolutions and aspect ratios and can adeptly handle various styles. \footnote{Refer to the supplementary material for more qualitative results.}

\subsection{Human Ratings}
We selected 532 challenging and representative prompts from GenEval~\citep{ghosh2024geneval}, T2I-CompBench~\citep{huang2023t2i}, PickScore~\citep{kirstain2023pick}, and Parti-prompts~\citep{yu2022scaling} to construct a comprehensive human evaluation benchmark. This benchmark assesses generated images based on prompt adherence, coherence, and realism.

To facilitate method comparison, we use Midjourney v6~\citep{midjourneyv6} as a baseline. Models are compared pairwise with Midjourney v6. The win rate in Fig.~\ref{fig: human preference} shows that D-JEPA$\cdot$T2I outperforms the diffusion model SD3.0 medium~\citep{esser2024scaling} and performs comparably to Midjourney v6. Although D-JEPA$\cdot$T2I currently trails behind the FLUX series and DALL·E 3, it's notable that FLUX has 12B parameters, markedly larger than our model's 2.6B scale. As for DALL·E 3, being a closed-source commercial model, its exact parameter count is unknown but inferred to surpass 2.6B based on DALL·E 2's 4.2B.

%% file: table/genai.tex
\begin{table}[]

\resizebox{\linewidth}{!}{
\begin{tabular}{lcccccc}
\rowcolor[HTML]{EFEFEF} 
\multicolumn{7}{l}{\cellcolor[HTML]{EFEFEF}\textit{VQAScores on ``basic'' prompts}}                                                                                                        \\
\multicolumn{1}{l|}{Method}                                                             & Attribute & Scene  & Spatial & Action & \multicolumn{1}{c|}{Part}                         & Avg  \\ \hline
\multicolumn{1}{l|}{SD v2.1~\citep{rombach2022high}}                                    & 0.75      & 0.79   & 0.73    & 0.73   & \multicolumn{1}{c|}{0.71}                         & 0.75 \\
\multicolumn{1}{l|}{SD-XL Turbo~\citep{sauer2025adversarial}}                           & 0.81      & 0.82   & 0.78    & 0.79   & \multicolumn{1}{c|}{0.78}                         & 0.80 \\
\multicolumn{1}{l|}{SD-XL~\citep{podell2023sdxl}}                                       & 0.82      & 0.85   & 0.80    & 0.80   & \multicolumn{1}{c|}{0.81}                         & 0.82 \\
\multicolumn{1}{l|}{DeepFloyd-IF~\citep{deepfloydIF}}                                   & 0.82      & 0.83   & 0.80    & 0.81   & \multicolumn{1}{c|}{0.81}                         & 0.82 \\
\multicolumn{1}{l|}{SD3.0 medium~\citep{esser2024scaling}}                              & 0.88      & 0.88   & 0.88    & 0.87   & \multicolumn{1}{c|}{0.89}                         & 0.88 \\
\rowcolor[HTML]{ECF4FF} 
\multicolumn{1}{l|}{\cellcolor[HTML]{ECF4FF}D-JEPA$\cdot$T2I}                           & 0.84      & 0.86   & 0.86    & 0.85   & \multicolumn{1}{c|}{\cellcolor[HTML]{ECF4FF}0.82} & 0.84 \\
\rowcolor[HTML]{FFFFC7} 
\multicolumn{1}{l|}{\cellcolor[HTML]{FFFFC7}Midjourney v6~\citep{midjourneyv6}}         & 0.86      & 0.88   & 0.86    & 0.87   & \multicolumn{1}{c|}{\cellcolor[HTML]{FFFFC7}0.85} & 0.86 \\
\rowcolor[HTML]{FFFFC7} 
\multicolumn{1}{l|}{\cellcolor[HTML]{FFFFC7}DALL$\cdot$E 3~\citep{betker2023improving}} & 0.91      & 0.91   & 0.90    & 0.90   & \multicolumn{1}{c|}{\cellcolor[HTML]{FFFFC7}0.91} & 0.90 \\
\rowcolor[HTML]{EFEFEF} 
\multicolumn{7}{l}{\cellcolor[HTML]{EFEFEF}\textit{VQAScore on ``advanced'' prompts}}                                                                                                      \\
\multicolumn{1}{l|}{Method}                                                             & Count     & Differ & Compare & Negate & \multicolumn{1}{c|}{Universal}                    & Avg  \\ \hline
\multicolumn{1}{l|}{SD v2.1~\citep{rombach2022high}}                                    & 0.66      & 0.64   & 0.65    & 0.51   & \multicolumn{1}{c|}{0.63}                         & 0.60 \\
\multicolumn{1}{l|}{SD-XL Turbo~\citep{sauer2025adversarial}}                           & 0.71      & 0.68   & 0.69    & 0.52   & \multicolumn{1}{c|}{0.66}                         & 0.63 \\
\multicolumn{1}{l|}{SD-XL~\citep{podell2023sdxl}}                                       & 0.72      & 0.70   & 0.69    & 0.50   & \multicolumn{1}{c|}{0.67}                         & 0.63 \\
\multicolumn{1}{l|}{DeepFloyd-IF~\citep{deepfloydIF}}                                   & 0.70      & 0.70   & 0.71    & 0.50   & \multicolumn{1}{c|}{0.65}                         & 0.63 \\
\multicolumn{1}{l|}{SD3.0 medium~\citep{esser2024scaling}}                              & 0.75      & 0.77   & 0.73    & 0.47   & \multicolumn{1}{c|}{0.70}                         & 0.65 \\
\rowcolor[HTML]{ECF4FF} 
\multicolumn{1}{l|}{\cellcolor[HTML]{ECF4FF}D-JEPA$\cdot$T2I}                           & 0.76      & 0.75   & 0.72    & 0.49   & \multicolumn{1}{c|}{\cellcolor[HTML]{ECF4FF}0.72} & 0.66 \\
\rowcolor[HTML]{FFFFC7} 
\multicolumn{1}{l|}{\cellcolor[HTML]{FFFFC7}Midjourney v6~\citep{midjourneyv6}}         & 0.77      & 0.77   & 0.76    & 0.50   & \multicolumn{1}{c|}{\cellcolor[HTML]{FFFFC7}0.73} & 0.68 \\
\rowcolor[HTML]{FFFFC7} 
\multicolumn{1}{l|}{\cellcolor[HTML]{FFFFC7}DALL$\cdot$E 3~\citep{betker2023improving}} & 0.80      & 0.80   & 0.77    & 0.49   & \multicolumn{1}{c|}{\cellcolor[HTML]{FFFFC7}0.75} & 0.69
\end{tabular}
}
\caption{VQAScores on ``basic'' and ``advanced'' prompts assessed by GenAI-Bench~\citep{li2024genai}.}
\label{tab: genai}
\vspace{-12pt}
\end{table}

%% file: figure/human_preference.tex
\begin{figure}
    \centering
    \includegraphics[width=\linewidth]{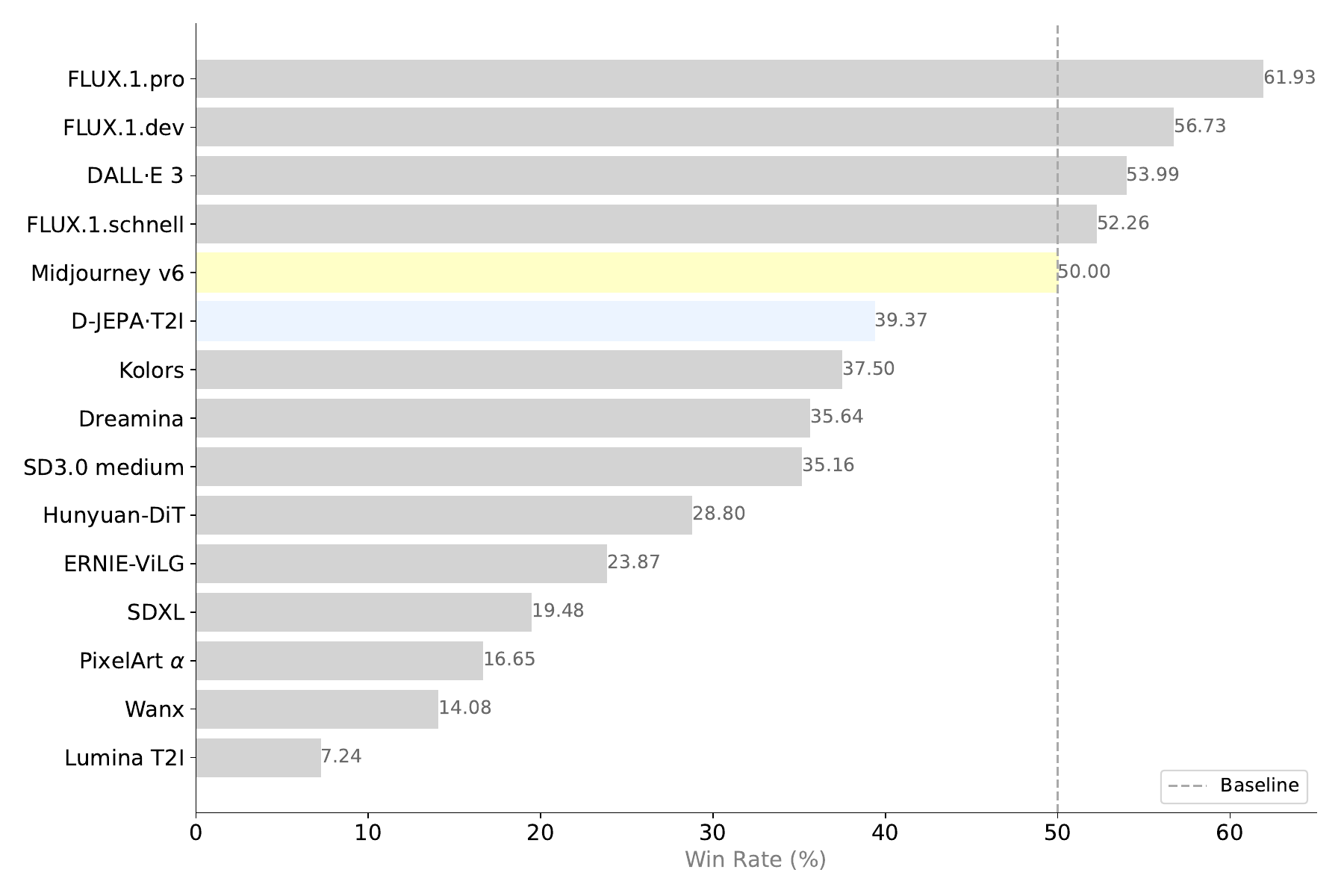}
    \caption{Models' win rate against Midjourney v6~\citep{midjourneyv6}. The evaluated models include Kolors~\citep{kolors}, Dreamina~\citep{jimengAI}, ERNIE-ViLG~\citep{feng2023ernie}, and Wanx~\citep{aliyun2023wanxiang}.}
    \label{fig: human preference}
    \vspace{-12pt}
\end{figure}

%% file: table/geneval_compbench_short.tex
\begin{table*}[htbp]
\centering

\resizebox{\textwidth}{!}{
\begin{tabular}{lcccccccccccc}
 &
   &
  \multicolumn{1}{c|}{} &
  \multicolumn{7}{c|}{GenEval} &
  \multicolumn{3}{c}{T2I-CompBench++} \\
\multicolumn{1}{l|}{Model} &
  NTP &
  \multicolumn{1}{c|}{\#Params} &
  \multicolumn{1}{c|}{Overall} &
  Single Obj. &
  Two Obj. &
  Counting &
  Colors &
  Position &
  \multicolumn{1}{c|}{Color Attr.} &
  Color &
  Shape &
  Texture \\ \hline
\multicolumn{1}{l|}{SD 3.0~\citep{esser2024scaling}} &
  $\times$ &
  \multicolumn{1}{c|}{2.0B} &
  \multicolumn{1}{c|}{0.62} &
  0.98 &
  0.74 &
  0.63 &
  0.67 &
  0.34 &
  \multicolumn{1}{c|}{0.36} &
  0.8132 &
  0.5885 &
  0.7334 \\
\rowcolor[HTML]{ECF4FF} 
\multicolumn{1}{l|}{\cellcolor[HTML]{ECF4FF}D-JEPA$\cdot$T2I} &
  $\checkmark$ &
  \multicolumn{1}{c|}{\cellcolor[HTML]{ECF4FF}2.6B} &
  \multicolumn{1}{c|}{\cellcolor[HTML]{ECF4FF}0.66} &
  0.99 &
  0.80 &
  0.59 &
  0.87 &
  0.22 &
  \multicolumn{1}{c|}{\cellcolor[HTML]{ECF4FF}0.47} &
  0.7585 &
  0.5036 &
  0.6355 \\
\multicolumn{1}{l|}{SDXL~\citep{podell2023sdxl}} &
  $\times$ &
  \multicolumn{1}{c|}{2.6B} &
  \multicolumn{1}{c|}{0.55} &
  0.98 &
  0.74 &
  0.39 &
  0.85 &
  0.15 &
  \multicolumn{1}{c|}{0.23} &
  0.5879 &
  0.4687 &
  0.5299 \\
\multicolumn{1}{l|}{LlamaGen~\citep{sun2024autoregressive}} &
  $\checkmark$ &
  \multicolumn{1}{c|}{3.1B} &
  \multicolumn{1}{c|}{0.32} &
  0.71 &
  0.34 &
  0.21 &
  0.58 &
  0.07 &
  \multicolumn{1}{c|}{0.04} &
  - &
  - &
  - \\
\multicolumn{1}{l|}{SD 3.0~\citep{esser2024scaling}} &
  $\times$ &
  \multicolumn{1}{c|}{4.0B} &
  \multicolumn{1}{c|}{0.64} &
  0.96 &
  0.80 &
  0.65 &
  0.73 &
  0.33 &
  \multicolumn{1}{c|}{0.37} &
  - &
  - &
  - \\
\multicolumn{1}{l|}{SD 3.0~\citep{esser2024scaling}} &
  $\times$ &
  \multicolumn{1}{c|}{8.0B} &
  \multicolumn{1}{c|}{0.68} &
  0.98 &
  0.84 &
  0.66 &
  0.74 &
  0.40 &
  \multicolumn{1}{c|}{0.43} &
  - &
  - &
  - \\
\multicolumn{1}{l|}{Emu3~\citep{wang2024emu3}} &
  $\checkmark$ &
  \multicolumn{1}{c|}{8.0B} &
  \multicolumn{1}{c|}{0.54} &
  0.98 &
  0.71 &
  0.34 &
  0.81 &
  0.17 &
  \multicolumn{1}{c|}{0.21} &
  - &
  - &
  - \\
\multicolumn{1}{l|}{Fluid~\citep{fan2024fluid}} &
  $\checkmark$ &
  \multicolumn{1}{c|}{10.5B} &
  \multicolumn{1}{c|}{0.69} &
  0.96 &
  0.83 &
  0.63 &
  0.80 &
  0.39 &
  \multicolumn{1}{c|}{0.51} &
  - &
  - &
  - \\
\rowcolor[HTML]{FFFFC7} 
\multicolumn{1}{l|}{\cellcolor[HTML]{FFFFC7}DALL$\cdot$E 3~\citep{betker2023improving}} &
  $\times$ &
  \multicolumn{1}{c|}{\cellcolor[HTML]{FFFFC7}-} &
  \multicolumn{1}{c|}{\cellcolor[HTML]{FFFFC7}0.67} &
  0.96 &
  0.87 &
  0.47 &
  0.83 &
  0.43 &
  \multicolumn{1}{c|}{\cellcolor[HTML]{FFFFC7}0.45} &
  0.7785 &
  0.6205 &
  0.7036 \\
\rowcolor[HTML]{FFFFC7} 
\multicolumn{1}{l|}{\cellcolor[HTML]{FFFFC7}Midjourney v6~\citep{midjourneyv6}} &
  $\times$ &
  \multicolumn{1}{c|}{\cellcolor[HTML]{FFFFC7}-} &
  \multicolumn{1}{c|}{\cellcolor[HTML]{FFFFC7}0.63} &
  0.96 &
  0.81 &
  0.56 &
  0.83 &
  0.22 &
  \multicolumn{1}{c|}{\cellcolor[HTML]{FFFFC7}0.42} &
  0.7503 &
  0.6885 &
  0.6101
\end{tabular}
}
\caption{Comprehensive comparison with state-of-the-art models on the GenEval~\citep{ghosh2024geneval} and T2I CompBench++~\citep{huang2023t2i} benchmarks. All listed metrics are obtained without employing DPO or prompt rewriting techniques. The symbol $\checkmark$ denotes the use of the Next Token Prediction (NTP) strategy for image sampling. DALL$\cdot$E 3~\citep{betker2023improving} and Midjourney v6~\citep{midjourneyv6} are commercial closed-source models, and their exact model sizes are not publicly available. A complete table is provided in the supplementary material.}
\label{tab: geneval compbench short}
\vspace{-12pt}
\end{table*}

%% file: sec/con.tex
\section{Conclusion}

In this work, we introduce D-JEPA$\cdot$T2I, along with the VoPE positional encoding and a data feedback training strategy, aimed at enhancing the capability of autoregressive models to generate high-resolution images. Our empirical results demonstrate comprehensive outperformance over both diffusion models and existing autoregressive models in the text-to-image generation task. Future research will delve into further validation of scaling laws on D-JEPA$\cdot$T2I and investigate the effective application of the D-JEPA architecture to video generation, as well as building a unified multimodal model.

%% file: sec/X_suppl.tex
\clearpage
\setcounter{page}{1}
\maketitlesupplementary

\section{Additional Figures and Tables}

\input{figure/teaser}
\paragraph{Teaser.} Fig.~\ref{fig: teaser} presents the complete teaser image.

\input{figure/dist}

\paragraph{Distribution.} Fig.~\ref{fig: dist} illustrates the issue of uneven natural data distribution.

\input{figure/djepa_t2i}
\paragraph{D-JEPA$\cdot$T2I.} Fig.~\ref{fig: djepa t2i} provides a detailed overview of the full network architecture.

\input{table/geneval_compbench}
\paragraph{Full Table.} Tab.~\ref{tab: geneval compbench} lists the complete comparison results.

\input{sec/sup/related}
\input{sec/sup/data_proc}
\input{sec/sup/exp_details}
\input{sec/sup/human_eval}

\input{sec/sup/ablation}

\input{sec/sup/comparison}

\input{sec/sup/limitation}

\input{sec/sup/prompt}

%% file: figure/teaser.tex
\begin{figure*}[ht]
    \centering
    \includegraphics[width=\linewidth]{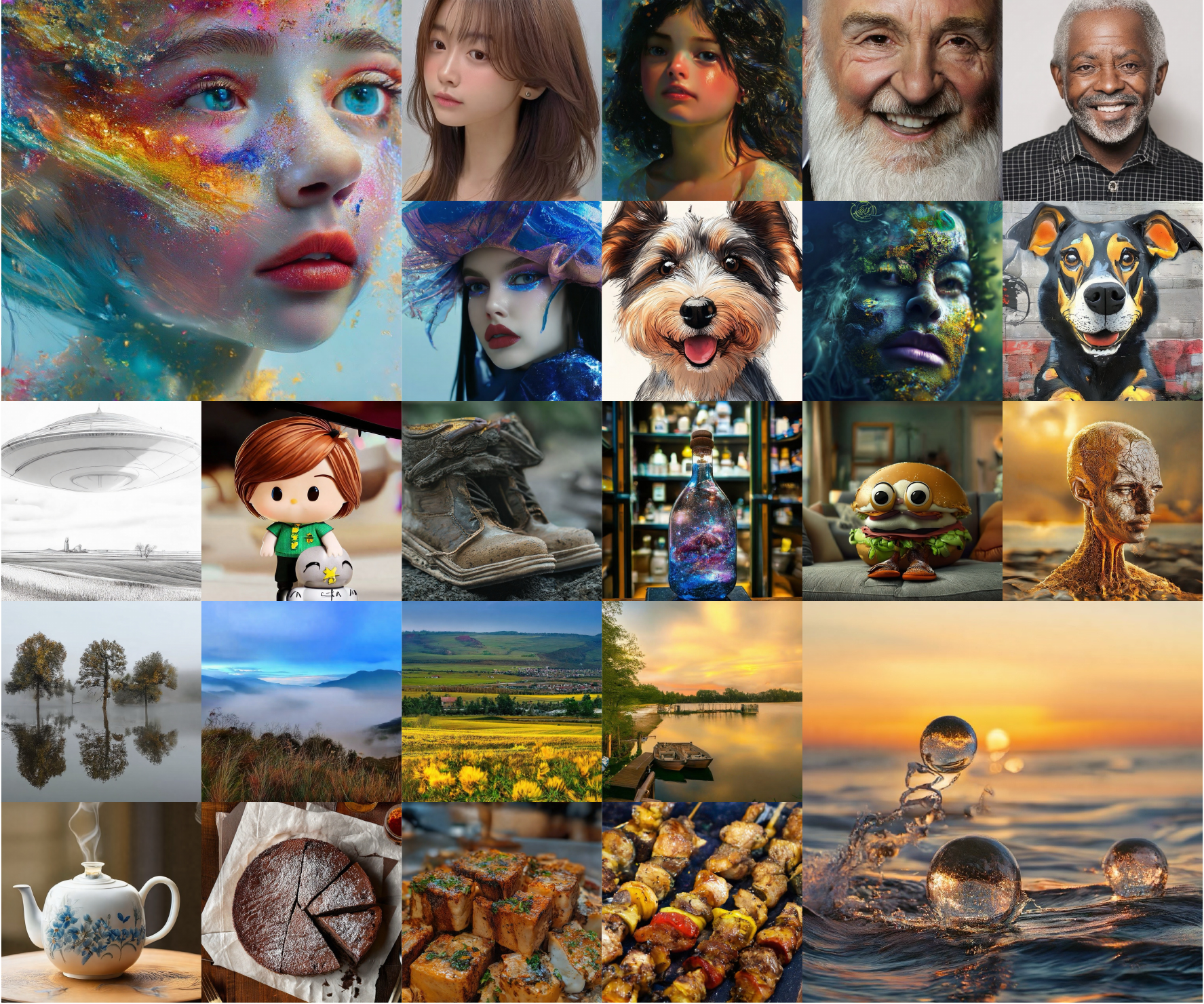}
    \caption{1K-resolution images sampled from D-JEPA$\cdot$T2I, showcasing its ability to generate high-fidelity, high-resolution images. (Best viewed when zoomed in.)}
    \label{fig: teaser}
    \vspace{-12pt}
\end{figure*}

%% file: figure/dist.tex
\begin{figure*}[ht]
    \centering
    \begin{subfigure}[b]{0.3\linewidth}
        \centering
        \includegraphics[width=\linewidth]{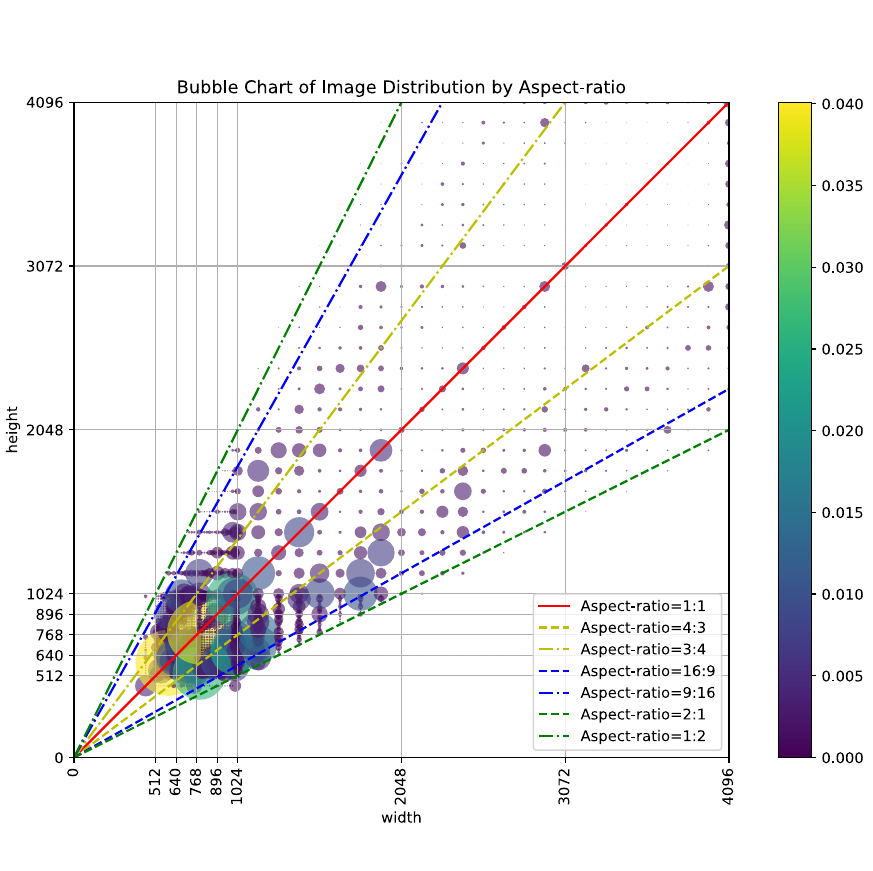}
    \end{subfigure}
    \hfill
    \begin{subfigure}[b]{0.3\linewidth}
        \centering
        \includegraphics[width=\linewidth]{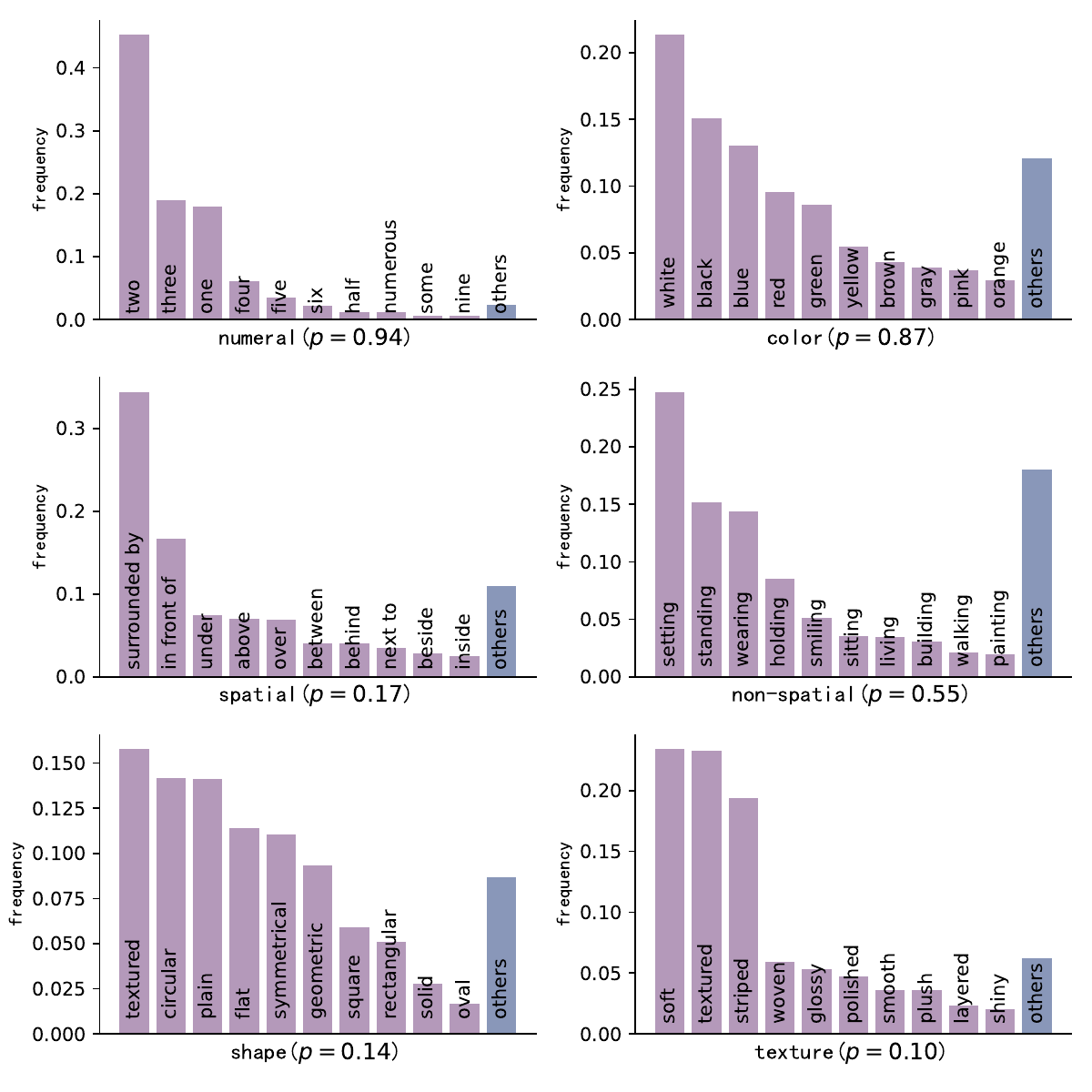}
    \end{subfigure}
    \hfill
    \begin{subfigure}[b]{0.3\linewidth}
        \centering
        \includegraphics[width=\linewidth]{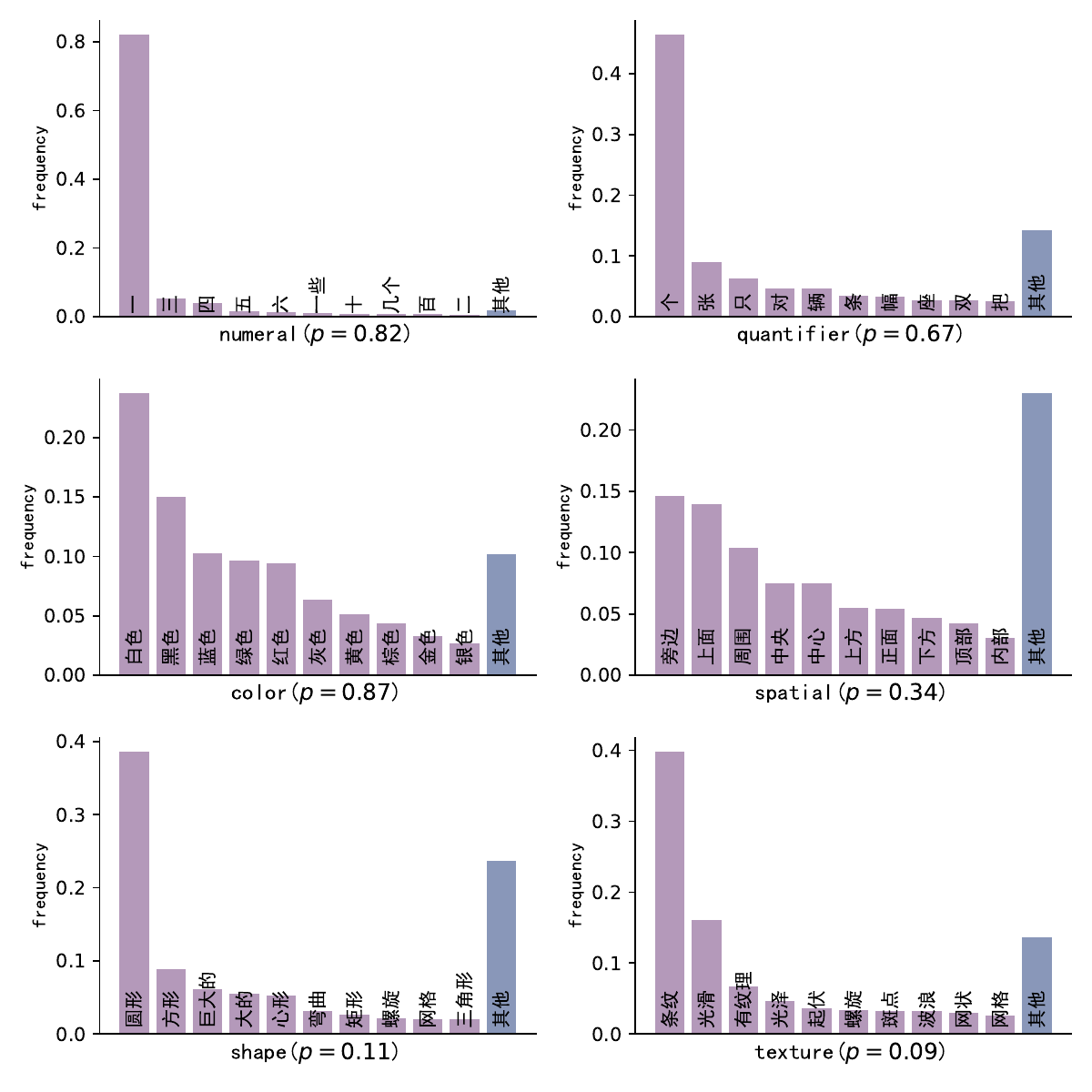}
    \end{subfigure}
    \caption{The resolution and notion distribution in the curated LAION dataset~\citep{schuhmann2022laion}, where $p$ represents the proportion of prompts containing a given notion. Additionally, we plot the distribution of translated Chinese characters in the third subplot as a supplementary reference, highlighting that the observed distribution imbalance is inherent to the dataset itself rather than being language-dependent.
    }
    \label{fig: dist}
    \vspace{-12pt}
\end{figure*}

%% file: figure/djepa_t2i.tex
\begin{figure*}
    \centering
    \begin{subfigure}{0.5\textwidth}
        \centering
        \includegraphics[width=\textwidth]{figure/djepa_t2i_img.pdf}
        \caption{Overview of all components.}
        \label{fig: framework}
    \end{subfigure}\hfill
    \begin{subfigure}{0.5\textwidth}
        \centering
        \includegraphics[width=\textwidth]{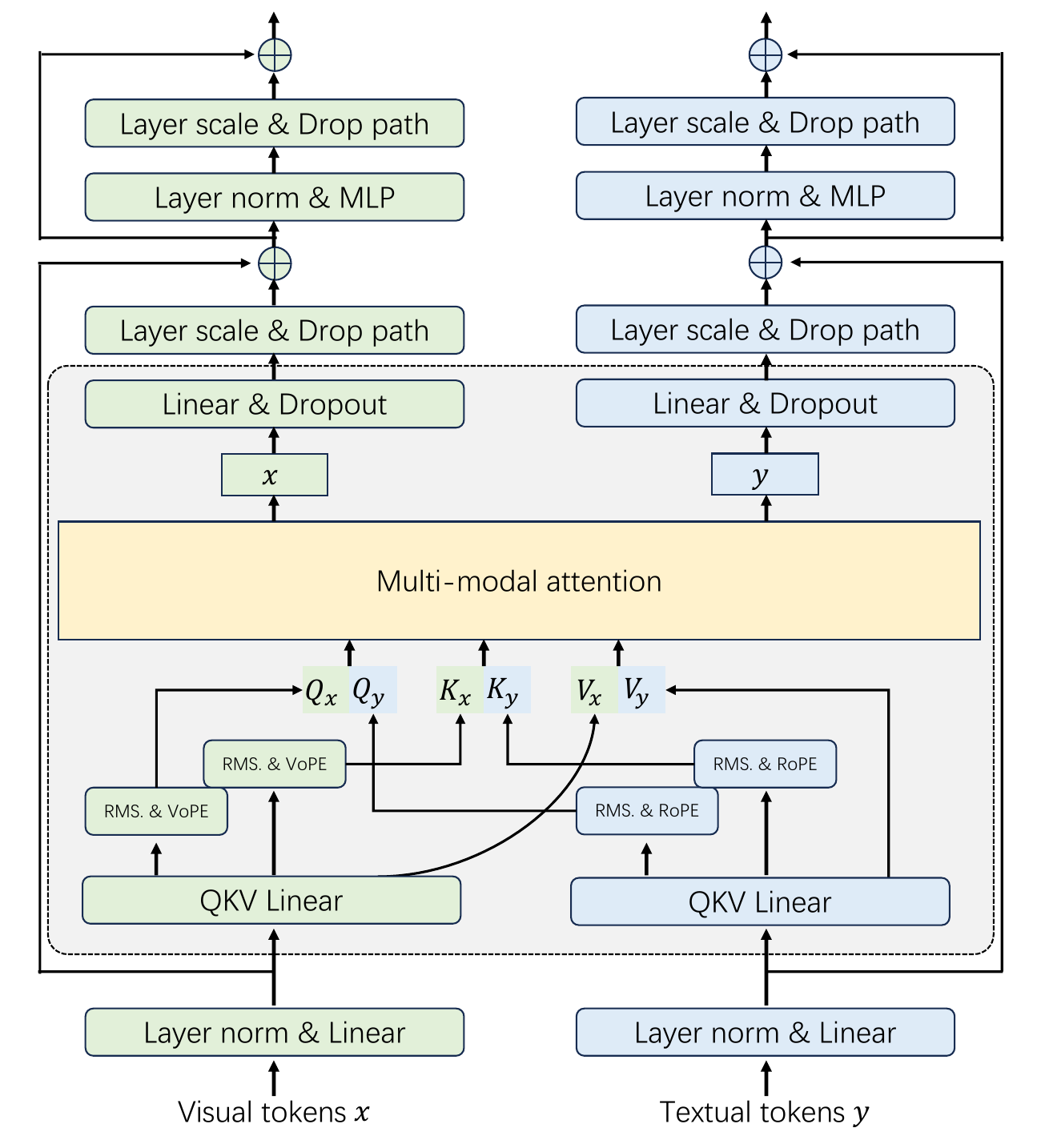}
        \caption{Multi-modal visual transformer block.}
        \label{fig: mmvit}
    \end{subfigure}
    \caption{Denoising with a Joint-Embedding Predictive Architecture for text-to-image synthesis. We present the overview of D-JEPA$\cdot$T2I in Fig.~\ref{fig: framework}. We employ T5-XXL~\citep{chung2024scaling} as the text encoder, and the KL-VAE pretrained by \citet{esser2024scaling} as the image encoder. Both textual and visual tokens are trimmed to no more than 256 and $256^2$ tokens for efficient training, respectively. The feature predictor $\gamma$, the context encoder $\phi$, and the target encoder $\bar{\phi}$ share the same network architecture, each consisting of several multimodal visual transformer blocks, as illustrated in Fig.~\ref{fig: mmvit}. The gradient is detached from the output of the target encoder $\bar{\phi}$, ensuring that it is only updated via Exponential Moving Average (EMA). Both the prediction loss $\mathcal{L}_{\text{pred}}$ and the flow matching loss $\mathcal{L}_{\text{flow}}$ are computed only for the masked visual tokens, following \citet{chen2024denoising}.}
    \label{fig: djepa t2i}
    \vspace{-12pt}
\end{figure*}

%% file: table/geneval_compbench.tex
\begin{table*}[htbp]
\centering

\resizebox{\textwidth}{!}{
\begin{tabular}{lcccccccccccc}
 &
   &
  \multicolumn{1}{c|}{} &
  \multicolumn{7}{c|}{GenEval} &
  \multicolumn{3}{c}{T2I-CompBench++} \\
\multicolumn{1}{l|}{Model} &
  NTP &
  \multicolumn{1}{c|}{\#Params} &
  \multicolumn{1}{c|}{Overall} &
  Single Obj. &
  Two Obj. &
  Counting &
  Colors &
  Position &
  \multicolumn{1}{c|}{Color Attr.} &
  Color &
  Shape &
  Texture \\ \hline
\rowcolor[HTML]{EFEFEF} 
\multicolumn{13}{l}{\cellcolor[HTML]{EFEFEF}\textit{Small model size}} \\
\multicolumn{1}{l|}{PixArt-$\alpha$~\citep{chen2023pixart}} &
  $\times$ &
  \multicolumn{1}{c|}{0.6B} &
  \multicolumn{1}{c|}{0.48} &
  0.98 &
  0.50 &
  0.44 &
  0.80 &
  0.08 &
  \multicolumn{1}{c|}{0.07} &
  0.4232 &
  0.3764 &
  0.4808 \\
\multicolumn{1}{l|}{SD v1.x~\citep{rombach2022high}} &
  $\times$ &
  \multicolumn{1}{c|}{0.9B} &
  \multicolumn{1}{c|}{0.43} &
  0.97 &
  0.38 &
  0.35 &
  0.76 &
  0.04 &
  \multicolumn{1}{c|}{0.06} &
  0.3765 &
  0.3576 &
  0.4156 \\
\multicolumn{1}{l|}{SD v2.x~\citep{rombach2022high}} &
  $\times$ &
  \multicolumn{1}{c|}{0.9B} &
  \multicolumn{1}{c|}{0.50} &
  0.98 &
  0.51 &
  0.44 &
  0.85 &
  0.07 &
  \multicolumn{1}{c|}{0.17} &
  0.5065 &
  0.4221 &
  0.4922 \\
\multicolumn{1}{l|}{SD 3.0~\citep{esser2024scaling}} &
  $\times$ &
  \multicolumn{1}{c|}{1.0B} &
  \multicolumn{1}{c|}{0.58} &
  0.97 &
  0.72 &
  0.52 &
  0.78 &
  0.16 &
  \multicolumn{1}{c|}{0.34} &
  - &
  - &
  - \\
\multicolumn{1}{l|}{Show-o~\citep{xie2024show}} &
  $\checkmark$ &
  \multicolumn{1}{c|}{1.3B} &
  \multicolumn{1}{c|}{0.53} &
  0.95 &
  0.52 &
  0.49 &
  0.82 &
  0.11 &
  \multicolumn{1}{c|}{0.28} &
  - &
  - &
  - \\
\multicolumn{1}{l|}{LDM~\citep{rombach2022high}} &
  $\times$ &
  \multicolumn{1}{c|}{1.4B} &
  \multicolumn{1}{c|}{0.37} &
  0.92 &
  0.29 &
  0.23 &
  0.70 &
  0.02 &
  \multicolumn{1}{c|}{0.05} &
  - &
  - &
  - \\
\multicolumn{1}{l|}{Hunyuan-DiT~\citep{li2024hunyuan}} &
  $\times$ &
  \multicolumn{1}{c|}{1.5B} &
  \multicolumn{1}{c|}{0.57} &
  0.96 &
  0.67 &
  0.59 &
  0.83 &
  0.11 &
  \multicolumn{1}{c|}{0.26} &
  0.6565 &
  0.3577 &
  0.4718 \\
\rowcolor[HTML]{EFEFEF} 
\multicolumn{13}{l}{\cellcolor[HTML]{EFEFEF}\textit{Mainstream model size}} \\
\multicolumn{1}{l|}{Lumina-T2I~\citep{zhuo2024lumina}} &
  $\times$ &
  \multicolumn{1}{c|}{2.0B} &
  \multicolumn{1}{c|}{0.39} &
  0.88 &
  0.34 &
  0.31 &
  0.67 &
  0.05 &
  \multicolumn{1}{c|}{0.09} &
  0.4081 &
  0.3008 &
  0.4071 \\
\multicolumn{1}{l|}{SD 3.0~\citep{esser2024scaling}} &
  $\times$ &
  \multicolumn{1}{c|}{2.0B} &
  \multicolumn{1}{c|}{0.62} &
  0.98 &
  0.74 &
  0.63 &
  0.67 &
  0.34 &
  \multicolumn{1}{c|}{0.36} &
  0.8132 &
  0.5885 &
  0.7334 \\
\rowcolor[HTML]{ECF4FF} 
\multicolumn{1}{l|}{\cellcolor[HTML]{ECF4FF}D-JEPA$\cdot$T2I} &
  $\checkmark$ &
  \multicolumn{1}{c|}{\cellcolor[HTML]{ECF4FF}2.6B} &
  \multicolumn{1}{c|}{\cellcolor[HTML]{ECF4FF}0.66} &
  0.99 &
  0.80 &
  0.59 &
  0.87 &
  0.22 &
  \multicolumn{1}{c|}{\cellcolor[HTML]{ECF4FF}0.47} &
  0.7585 &
  0.5036 &
  0.6355 \\
\multicolumn{1}{l|}{SDXL~\citep{podell2023sdxl}} &
  $\times$ &
  \multicolumn{1}{c|}{2.6B} &
  \multicolumn{1}{c|}{0.55} &
  0.98 &
  0.74 &
  0.39 &
  0.85 &
  0.15 &
  \multicolumn{1}{c|}{0.23} &
  0.5879 &
  0.4687 &
  0.5299 \\
\multicolumn{1}{l|}{LlamaGen~\citep{sun2024autoregressive}} &
  $\checkmark$ &
  \multicolumn{1}{c|}{3.1B} &
  \multicolumn{1}{c|}{0.32} &
  0.71 &
  0.34 &
  0.21 &
  0.58 &
  0.07 &
  \multicolumn{1}{c|}{0.04} &
  - &
  - &
  - \\
\multicolumn{1}{l|}{SD 3.0~\citep{esser2024scaling}} &
  $\times$ &
  \multicolumn{1}{c|}{4.0B} &
  \multicolumn{1}{c|}{0.64} &
  0.96 &
  0.80 &
  0.65 &
  0.73 &
  0.33 &
  \multicolumn{1}{c|}{0.37} &
  - &
  - &
  - \\
\multicolumn{1}{l|}{DALL$\cdot$E 2~\citep{ramesh2022hierarchical}} &
  $\times$ &
  \multicolumn{1}{c|}{4.2B} &
  \multicolumn{1}{c|}{0.52} &
  0.94 &
  0.66 &
  0.49 &
  0.77 &
  0.10 &
  \multicolumn{1}{c|}{0.19} &
  - &
  - &
  - \\
\rowcolor[HTML]{EFEFEF} 
\multicolumn{13}{l}{\cellcolor[HTML]{EFEFEF}\textit{Extensive model size}} \\
\multicolumn{1}{l|}{Chameleon~\citep{team2024chameleon}} &
  $\checkmark$ &
  \multicolumn{1}{c|}{7.0B} &
  \multicolumn{1}{c|}{0.39} &
  - &
  - &
  - &
  - &
  - &
  \multicolumn{1}{c|}{-} &
  - &
  - &
  - \\
\multicolumn{1}{l|}{Transfusion~\citep{zhou2024transfusion}} &
  $\checkmark$ &
  \multicolumn{1}{c|}{7.3B} &
  \multicolumn{1}{c|}{0.63} &
  - &
  - &
  - &
  - &
  - &
  \multicolumn{1}{c|}{-} &
  - &
  - &
  - \\
\multicolumn{1}{l|}{SD 3.0~\citep{esser2024scaling}} &
  $\times$ &
  \multicolumn{1}{c|}{8.0B} &
  \multicolumn{1}{c|}{0.68} &
  0.98 &
  0.84 &
  0.66 &
  0.74 &
  0.40 &
  \multicolumn{1}{c|}{0.43} &
  - &
  - &
  - \\
\multicolumn{1}{l|}{Emu3~\citep{wang2024emu3}} &
  $\checkmark$ &
  \multicolumn{1}{c|}{8.0B} &
  \multicolumn{1}{c|}{0.54} &
  0.98 &
  0.71 &
  0.34 &
  0.81 &
  0.17 &
  \multicolumn{1}{c|}{0.21} &
  - &
  - &
  - \\
\multicolumn{1}{l|}{Fluid~\citep{fan2024fluid}} &
  $\checkmark$ &
  \multicolumn{1}{c|}{10.5B} &
  \multicolumn{1}{c|}{0.69} &
  0.96 &
  0.83 &
  0.63 &
  0.80 &
  0.39 &
  \multicolumn{1}{c|}{0.51} &
  - &
  - &
  - \\
\multicolumn{1}{l|}{FLUX.1.dev~\citep{flux1}} &
  $\times$ &
  \multicolumn{1}{c|}{12.0B} &
  \multicolumn{1}{c|}{-} &
  - &
  - &
  - &
  - &
  - &
  \multicolumn{1}{c|}{-} &
  0.7407 &
  0.5718 &
  0.6922 \\
\rowcolor[HTML]{FFFFC7} 
\multicolumn{1}{l|}{\cellcolor[HTML]{FFFFC7}DALL$\cdot$E 3~\citep{betker2023improving}} &
  $\times$ &
  \multicolumn{1}{c|}{\cellcolor[HTML]{FFFFC7}-} &
  \multicolumn{1}{c|}{\cellcolor[HTML]{FFFFC7}0.67} &
  0.96 &
  0.87 &
  0.47 &
  0.83 &
  0.43 &
  \multicolumn{1}{c|}{\cellcolor[HTML]{FFFFC7}0.45} &
  0.7785 &
  0.6205 &
  0.7036 \\
\rowcolor[HTML]{FFFFC7} 
\multicolumn{1}{l|}{\cellcolor[HTML]{FFFFC7}Midjourney v6~\citep{midjourneyv6}} &
  $\times$ &
  \multicolumn{1}{c|}{\cellcolor[HTML]{FFFFC7}-} &
  \multicolumn{1}{c|}{\cellcolor[HTML]{FFFFC7}0.63} &
  0.96 &
  0.81 &
  0.56 &
  0.83 &
  0.22 &
  \multicolumn{1}{c|}{\cellcolor[HTML]{FFFFC7}0.42} &
  0.7503 &
  0.6885 &
  0.6101
\end{tabular}
}
\caption{Comprehensive comparison with state-of-the-art models on the GenEval~\citep{ghosh2024geneval} and T2I CompBench++~\citep{huang2023t2i} benchmarks. All listed metrics are obtained without employing DPO and prompt rewriting techniques. The symbol $\checkmark$ denotes the use of the Next Token Prediction (NTP) strategy for image sampling. DALL$\cdot$E 3~\citep{betker2023improving} and Midjourney v6~\citep{midjourneyv6} are commercial closed-source models, and we do not have access to their exact model sizes.}
\label{tab: geneval compbench}
\vspace{-12pt}
\end{table*}

%% file: sec/sup/related.tex
\section{Related Work}

\paragraph{Diffusion Models.} 

The pioneering work of \citet{sohl2015deep, songscore, ho2020denoising} established a methodology for data generation by approximating the reverse ordinary differential equation of a stochastic forward process that transforms data into noise. This novel approach has become a cornerstone in both image~\citep{dhariwal2021diffusion, ramesh2022hierarchical, saharia2022photorealistic, rombach2022high, balaji2022ediffi} and video generation domains~\citep{singer2022makeavideo, ho2022imagen, esser2023structure, blattmann2023align, gupta2023photorealistic}. 

Expanding on derivations based on the variational lower bound on negative likelihood~\citep{sohl2015deep} and score matching~\citep{hyvarinen2005estimation, vincent2011connection, song2020generative}, researchers have explored various formulations of forward and reverse processes~\citep{songscore, dockhorn2021score}, model parameterizations~\citep{ho2020denoising, ho2021classifier, karras2022elucidating}, loss weightings~\citep{ho2020denoising, karras2022elucidating}, and ODE solvers~\citep{song2022denoising, lu2023dpmsolver, dockhorn2022genie}. Significant contributions by \citet{kingma2024understanding} and \citet{karras2022elucidating} have provided unified formulations that offer novel theoretical and practical insights for both training~\citep{karras2022elucidating, kingma2024understanding} and inference~\citep{karras2022elucidating}. Nonetheless, common ODE trajectories often exhibit significant curvature~\citep{karras2022elucidating, liu2022flow}, requiring numerous solver steps and complicating rapid inference. 

Recent research efforts have aimed at improving learning and sampling methods~\citep{song2019generative, song2020denoising, lu2023dpmsolver, bao2022analytic}, utilizing guidance techniques~\citep{ho2021classifier, nichol2021glide}, leveraging latent learning~\citep{rombach2022high}, and advancing architectural designs~\citep{ho2022cascaded, peebles2023scalable, saharia2022photorealistic, xue2024raphael}. Innovative models like DiT~\citep{peebles2023scalable} and U-ViT~\citep{bao2023all} integrate or replace the U-Net with transformers, inspiring advancements in image~\citep{chen2023pixart, chen2024pixart} and video synthesis systems~\citep{bar2024lumiere, gupta2023photorealistic}, including Stable Diffusion 3.0~\citep{esser2024scaling}, SORA~\citep{brooks2024sora}, and Vidu~\citep{bao2024vidu}.

\paragraph{Rectified Flow Models.}

Rectified flow models~\citep{liu2022flow, albergo2022building, lipman2023flow} approach generative modeling by constructing a transport map between two distributions via an ordinary differential equation (ODE). This technique is closely related to continuous normalizing flows~(CNF)~\citep{chen2018neural} and diffusion models. Compared to CNFs, rectified flows and stochastic interpolants eliminate the need for ODE simulation during training. Compared to diffusion models, they can generate ODEs that are faster to simulate than the probability flow ODE~\citep{songscore} associated with diffusion models. However, rectified flow models do not yield optimal transport solutions, which has led to efforts aiming to minimize trajectory curvature~\citep{lee2023minimizing, alex2023improving, aramalex2023multisample}. Recent works~\citep{dao2023flow, ma2024sit} demonstrate the feasibility of rectified flow formulations for class-conditional image synthesis, while \citet{fischer2023boosting} employs them for latent-space upsampling, and \citet{liu2023instaflow} leverages the reflow procedure from \cite{liu2022flow} to distill a pretrained text-to-image model~\citep{rombach2022high}. In this study, we focus specifically on employing rectified flows as the flow matching loss to effectively model the distribution of each token.

\paragraph{Masked Prediction Models.}

MaskGIT~\citep{chang2022maskgit} utilizes a vector-quantized autoencoder~\citep{razavi2019generating} along with a masked prediction transformer similar to BERT~\citep{kenton2019bert, bao2021beit, he2022masked} to generate discrete tokens via a greedy algorithm. This approach is extended to videos by MagViT~\citep{yu2023magvit}, and further refined by MagViT-2~\citep{yu2023language}, which enhances these techniques~\citep{chang2022maskgit, yu2023magvit} with an improved VQ-VAE~\citep{razavi2019generating} for both images and videos. MUSE~\citep{chang2023muse} scales MaskGIT to 3 billion parameters. Recent developments by \citet{chen2024denoising, li2024autoregressive, tian2024visual} have shifted away from reliance on VQ-VAE, opting instead for KL-VAE, yielding better results. Despite these advancements, masked prediction models (or autoregressive models) still lag behind diffusion models and rectified flow models in the context of high-resolution text-to-image generation tasks.

%% file: sec/sup/data_proc.tex
\section{Data Processing}

To obtain high-quality image-text pairs and mitigate data bias, we performed a comprehensive analysis of the dataset to achieve a more balanced distribution of training data.

Beyond conventional filters such as resolution, OCR\citep{song2023mugs}, face detection\footnote{\url{https://huggingface.co/arnabdhar/YOLOv8-Face-Detection}}, aesthetic score\footnote{\url{https://github.com/LAION-AI/aesthetic-predictor}} and NSFW content\footnote{\url{https://github.com/LAION-AI/CLIP-based-NSFW-Detector}}, we examined the distribution of basic information for every data source and the distribution among different feature combinations, such as resolution and aspect ratio. This approach enabled us to identify the fundamental characteristics of the different datasets.

Our analysis focused on the distribution of entity words and attribute words in the captions obtained using InternVL2~\citep{chen2023internvl}. Utilizing standard NLP tools\footnote{\url{https://github.com/explosion/spaCy}}, we extracted entity words, including abstract concepts, from the captions. The attribute words primarily included numerals, quantifiers (specific to Chinese), spatial terms, non-spatial terms (specific to English), shapes, colors, and textures, in conjunction with metrics from T2I-CompBench~\citep{boomb0omT2IBenchmark} and GenEval~\citep{ghosh2024geneval}. Initially, we obtained a comprehensive vocabulary list for specific attributes using GPT-4~\citep{achiam2023gpt}. The prompt was: ``Please return as many words as possible that describe numerals/colors/shapes/positions/textures in English/Chinese. Directly return the list result [], DO NOT hallucinate, and avoid polite language.'' For non-spatial terms, we referred to the 1000 prompts provided in T2I-CompBench \footnote{\url{https://github.com/Karine-Huang/T2I-CompBench/blob/main/examples/dataset/non\_spatial.txt}} to analyze the gerunds. We then employed regular expressions to match these words in the captions and counted their frequency of occurrence, prioritizing longer attribute words during the matching process.

%% file: sec/sup/exp_details.tex
\section{Experiments Details}
\input{figure/norm}


\paragraph{Scaling to 4K Resolution.}

To support 4K resolution image generation, we need to resolve two main issues. First, high-resolution images can lead to insufficient GPU memory during VAE encoding and decoding. Second, the token sequences corresponding to high-resolution images are considerably lengthy, resulting in memory limitations during training. Common engineering solutions involve using Zero Redundancy Optimizer~\citep{rasley2020deepspeed} and Megatron~\footnote{https://github.com/NVIDIA/Megatron-LM} to offload some optimizer states to the CPU, or employing strategies like model parallelism and tensor parallelism.

Here, we devise a data-efficient training technique to enable training at 4K resolution and beyond with D-JEPA$\cdot$T2I. Unlike diffusion models, which require processing entire images per iteration, the autoregressive D-JEPA$\cdot$T2I can predict random portions of images based on random context tokens. Thus, when the total number of tokens exceeds 4096 (i.e., 256 tokens per side), we apply a random drop strategy to maintain the token count at 4096 or fewer. While training with a random subset of tokens might limit the model's ability to learn global features, we found that through dynamic resolution training, and leveraging the characteristics of VoPE, D-JEPA$\cdot$T2I can quickly adapt to higher resolution generation.

Regarding the VAE encoding and decoding processes, when the image resolution surpasses 2K, we employ a tiling strategy, dividing the image into four parts for encoding/decoding to reduce memory consumption. Although this method may cause discontinuities at the edges, it is currently acceptable.
\input{figure/4k}
Fig.~\ref{fig: 4k} illustrates the results of image generation across a range from 1K to 4K, demonstrating the potential of D-JEPA$\cdot$T2I for generating high-resolution images.

\paragraph{Inference Details.} Empirically, we found that no more than 128 autoregressive steps are sufficient to generate images of any aspect ratio and resolution for D-JEPA$\cdot$T2I. Therefore, unless otherwise specified, we use 128 autoregressive steps in our experiments. We observed that the classifier-free guidance (CFG)~\citep{ho2022classifier} significantly influences the quality of the generated images. A higher CFG often achieves better evaluation metrics on benchmarks like GenEval~\citep{ghosh2024geneval}, but may compromise aesthetic qualities. Thus, in practice, for the experiments shown in Tab.~\ref{tab: genai} and Tab.~\ref{tab: geneval compbench}, we set the CFG value to 6.0, while in other scenarios, we set it to 2.0. Additionally, we use a time shifting factor, as described in ~\cite{gao2024lumina}, to adjust the allocation of steps in the denoising process. Through grid search, we determined this value to be 4.5.

%% file: figure/norm.tex
\begin{figure}
    \centering
    \begin{subfigure}[b]{0.45\textwidth}
        \centering
        \includegraphics[width=\linewidth]{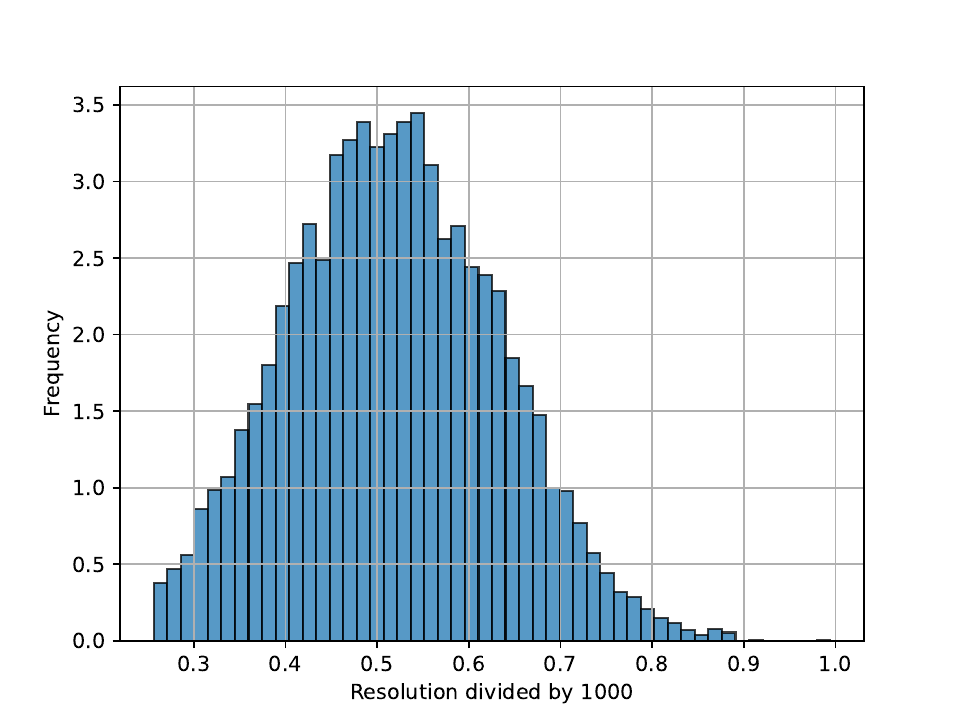}
    \end{subfigure}
    \hfill
    \begin{subfigure}[b]{0.45\textwidth}
        \centering
        \includegraphics[width=\linewidth]{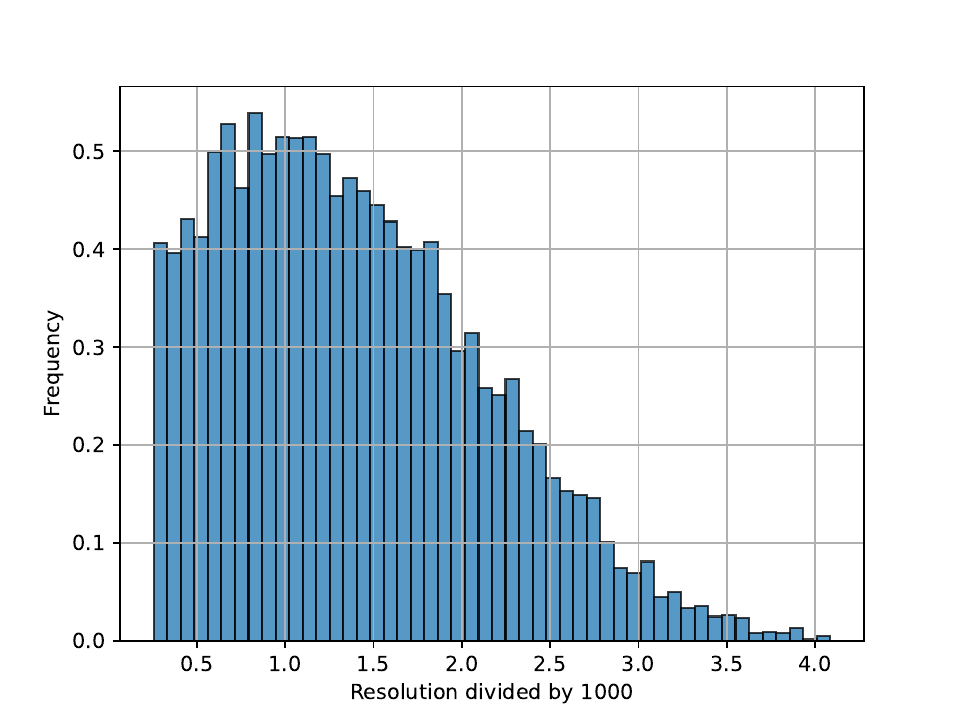}
    \end{subfigure}
    \caption{Dynamic resolution distribution during phase two training. Left: $\mathrm{trunc\_norm}(0.512, 0.12, 0.256, 1.024)$, primarily used to ensure the model can generate images with any aspect ratio and resolution within 1k resolution. Right: $\mathrm{trunc\_norm}(1.024, 1.0, 0.256, 4.096)$, used to extend the model's capability to generate images from 1k to 4k resolution.}
    \label{fig: norm}
    \vspace{-12pt}
\end{figure}

%% file: figure/4k.tex
\begin{figure*}
    \centering
    \includegraphics[width=0.8\linewidth]{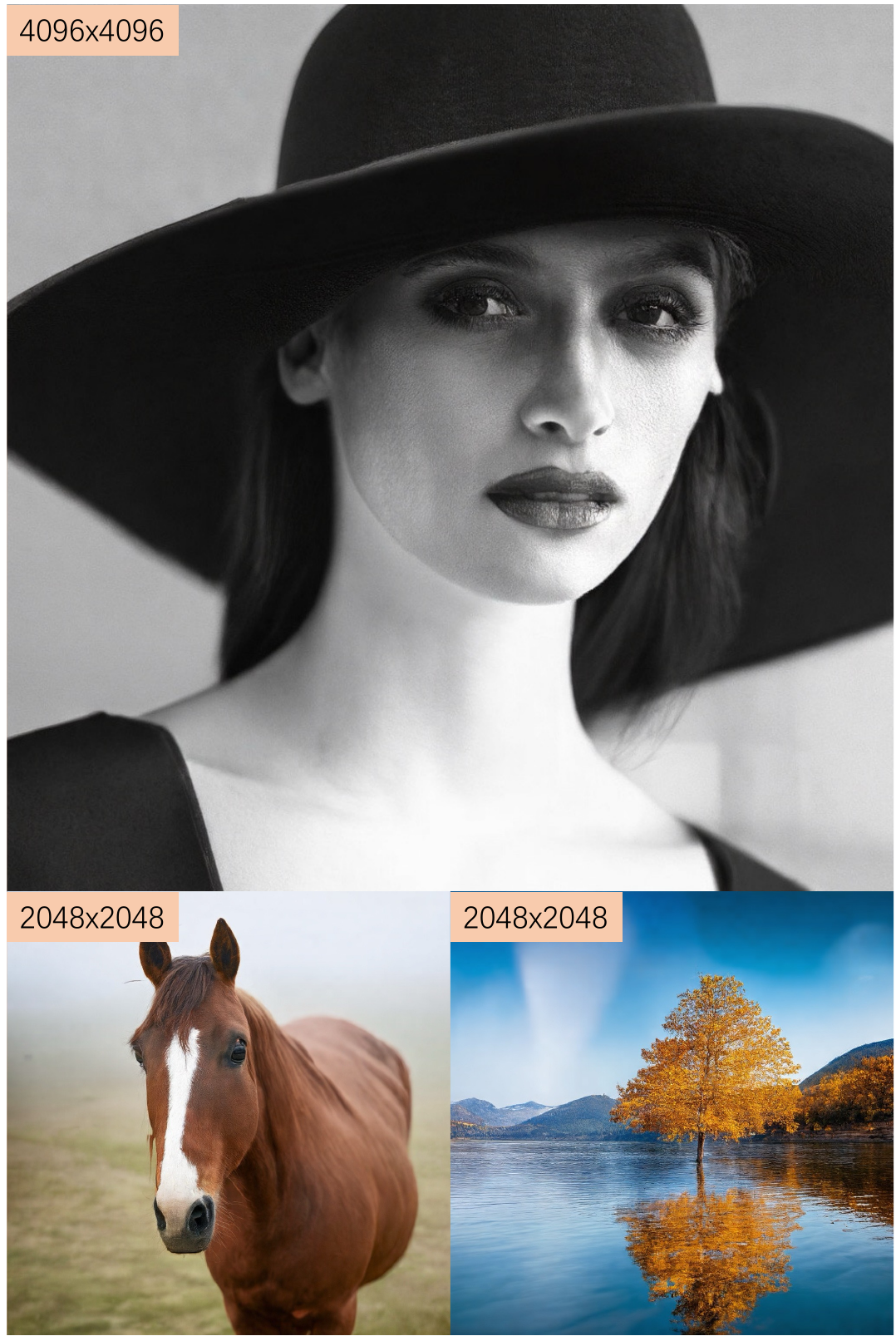}
    \caption{Ultra-high-resolution images generated by D-JEPA$\cdot$T2I.}
    \label{fig: 4k}
\end{figure*}

%% file: sec/sup/human_eval.tex
\section{Human Evaluation}

We curated a comprehensive human evaluation benchmark by selecting 532 challenging and representative prompts from GenEval~\citep{ghosh2024geneval}, T2I-CompBench~\citep{huang2023t2i}, PickScore~\citep{kirstain2023pick}, and Parti-prompts~\citep{yu2022scaling}. This benchmark assesses the generated images based on their adherence to prompts, coherence, and realism.

To evaluate human preferences across these categories, raters were presented with paired outputs from two models and asked the following questions:
\begin{itemize}
    \item \textit{Prompt Adherence}: Which image more accurately represents and faithfully follows the provided text?
    \item \textit{Coherence}: Which image better encapsulates the elements specified in the description?
    \item \textit{Realism}: Given the prompt, which image exhibits higher quality and realism?
\end{itemize}

For each question, raters assessed the images as ``better'', ``same'', or ``worse'' when comparing outputs from different models. The questions were prioritized with prompt adherence being most critical, followed by coherence, and then realism. If a "same" result was reached at one priority level, only then were results at the subsequent level considered.

The win rate for each model against Midjourney v6~\citep{midjourneyv6}, as shown in Fig.~\ref{fig: human preference}, is calculated using the formula:

$$
\text{win rate} = \frac{\# \text{better} + 0.5 \times \# \text{same}}{\# \text{better} + \# \text{same} + \# \text{worse}}.
$$

%% file: sec/sup/ablation.tex
\section{Ablation Study of Data Feedback}

\input{table/ablation}
\paragraph{Statistical Analysis Sampling Accelerates Early-Stage Model Convergence.}  
Data feedback plays a crucial role in improving the convergence speed of the model. By continuously analyzing the performance of generated images and adjusting the training dataset accordingly, the model learns more effectively from iterations that most significantly reduce errors and enhance accuracy. This targeted learning approach enables the model to converge faster than traditional static training methods. As shown in Tab.~\ref{tab: data feedback}, by comparing experiments AB and CD, we achieve an overall score of 0.50 within 100k steps on the GenEval~\citep{ghosh2024geneval} benchmark using statistical analysis sampling. In contrast, with random sampling, even after 200k steps, the model only reaches a score of 0.48. This results in an almost twofold improvement in convergence speed.  

\paragraph{Critic Model Sampling Effectively Mitigates Failure Cases in Late-Stage Training.}  
By comparing experiments DE and FG, we find that critic model sampling with automated annotation further accelerates model convergence. With this approach, we achieve an overall score of 0.63 on GenEval at around 200k training steps. In contrast, relying solely on statistical analysis sampling requires 300k iterations to reach the same performance.  

\input{figure/failure}  
Our experiments (HI) further demonstrate that incorporating critic model sampling with human annotation substantially reduces failure cases (illustrated in Fig.~\ref{fig: failure}) in the later stages of training. By dynamically adjusting the data distribution based on model performance, the critic model ensures a more balanced and robust learning process, leading to improved overall generation quality. Correspondingly, the model's win rate against Midjourney v6 significantly increases from 17.3\% to 35.6\%, underscoring the crucial role of data feedback in enhancing model performance.  

\section{Properties of VoPE}
\input{figure/amb_a}
\input{figure/amb_b}
\input{figure/amb_c}
\input{figure/amb_d}
\paragraph{Arbitrary Aspect Ratios and Continuous Resolutions Generation.} Due to VoPE's ability to consistently align all images to the reference grid size $g \times g$ during training, it ensures that images maintain consistency and plausibility in content when generating at any aspect ratio and continuous resolution, preventing distortions. This flexibility is vital for adapting the model to various use cases where specific image dimensions are needed. Fig.~\ref{fig: amb a}, ~\ref{fig: amb b}, ~\ref{fig: amb c} and ~\ref{fig: amb d} show some samples generated by D-JEPA$\cdot$T2I, demonstrating that regardless of extreme aspect ratios or resolutions, D-JEPA$\cdot$T2I can produce coherent, high-quality images.

\input{figure/bias}
\paragraph{Layout Control by Relative Positional Offset $b$.} During training, the pixel normalization operation involved in VoPE ensures that each image aligns to the center of the reference grid, achieved through the relative positional offset $b$. In the inference phase, we can manipulate the relative positional offset $b$ to control the layout of the generated image, particularly when the generated objects are off-center or the main subject is incomplete. Fig.~\ref{fig: bias} illustrates how adjusting the relative positional offset $b$ can alter the overall layout to produce more satisfactory results.

%% file: table/ablation.tex
\begin{table*}[h]
\centering
\begin{tabular}{c|c|ccccc}
Group & Steps$\sim$(k) & SAS & CMS(Auto) & CMS(Human)               & GenEval Overall & Win Rate(\%) \\ \hline
A     & $0\sim 100$    & w/o & w/o       & \multicolumn{1}{c|}{w/o} & 0.38            & -            \\
B     & $100 \sim 200$ & w/o & w/o       & \multicolumn{1}{c|}{w/o} & 0.48            & -            \\ \hline
C     & $0 \sim 100$   & w/  & w/o       & \multicolumn{1}{c|}{w/o} & 0.50            & -            \\
D     & $100 \sim 200$ & w/  & w/o       & \multicolumn{1}{c|}{w/o} & 0.60            & -            \\ \hline
E     & $200 \sim 300$ & w/  & w/o       & \multicolumn{1}{c|}{w/o} & 0.63            & -            \\
F     & $100\sim 200$  & w/  & w/        & \multicolumn{1}{c|}{w/o} & 0.63            & -            \\ \hline
G     & $200 \sim 300$ & w/  & w/        & \multicolumn{1}{c|}{w/o} & 0.66            & -            \\ \hline
H     & $300 \sim 600$ & w/  & w/        & \multicolumn{1}{c|}{w/o} & -               & 17.3         \\
I     & $300 \sim 600$ & w/  & w/o       & \multicolumn{1}{c|}{w/}  & -               & 35.6        
\end{tabular}
\caption{Ablation study on data feedback. SAS: Statistical Analysis Sampling, CMS~(Auto): Critic Model Sampling with automated annotation, CMS~(Human): Critic Model Sampling with human annotation. The reported win rate is measured against Midjourney V6.  
}
\label{tab: data feedback}
\end{table*}

%% file: figure/failure.tex
\begin{figure}
    \centering
    \includegraphics[width=\linewidth]{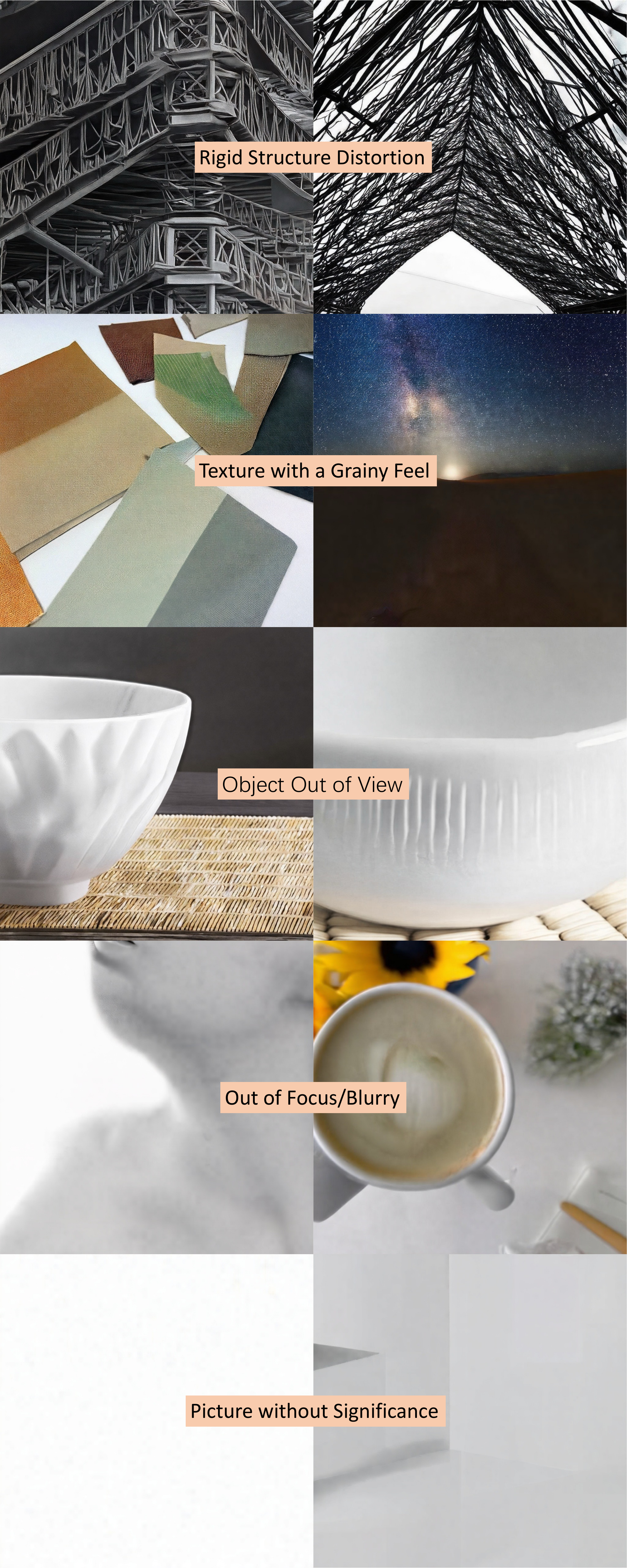}
    \caption{Common failure cases in image generation (without data feedback). Notably, training with data feedback can significantly reduce these failure cases.}
    \label{fig: failure}
\end{figure}

%% file: figure/amb_a.tex
\begin{figure*}
    \centering
    \includegraphics[width=\linewidth]{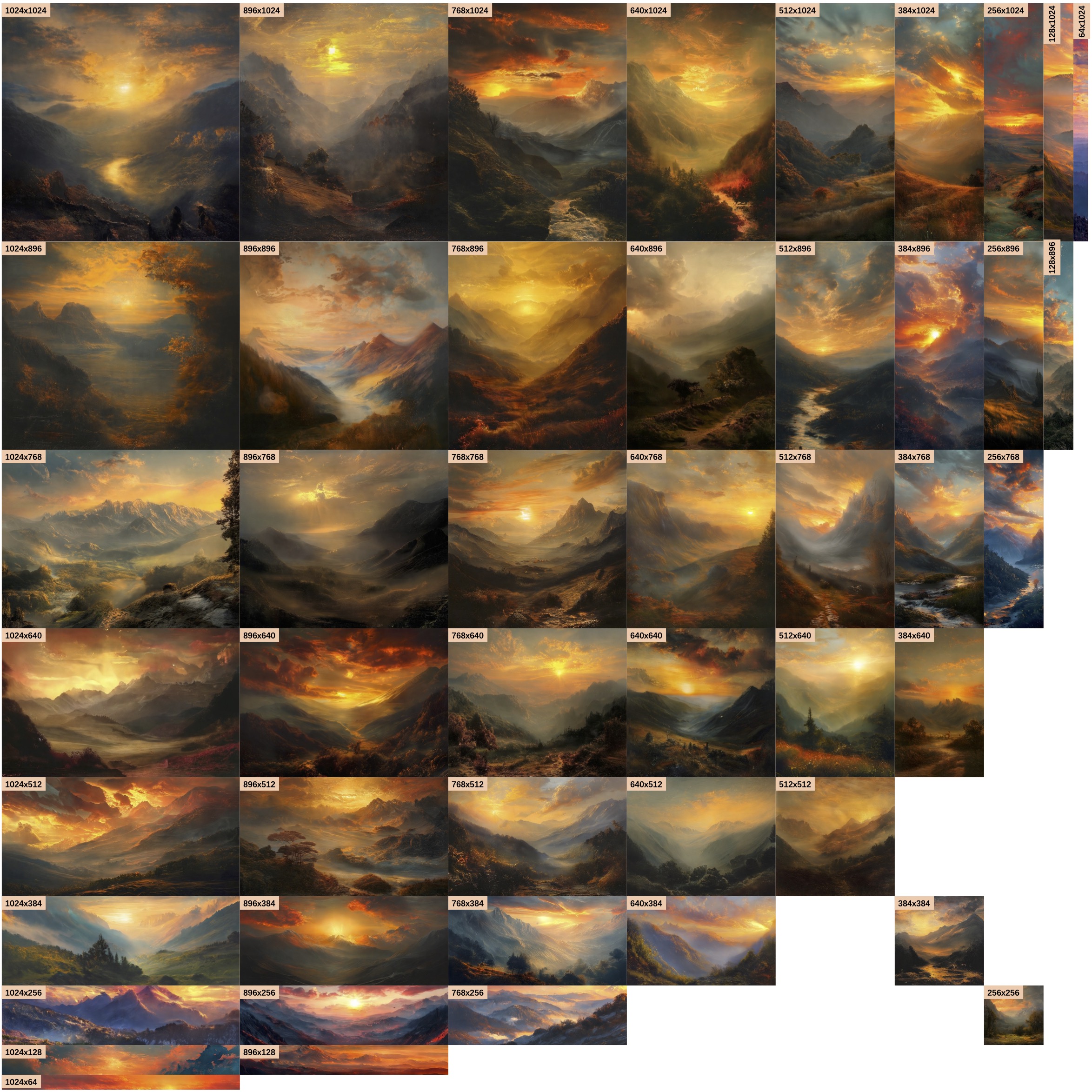}
    \caption{D-JEPA$\cdot$T2I can generate arbitrary aspect ratios and continuous resolutions with VoPE. Prompt: ``A gorgeous mountain landscape at sunset. Masterful painting by Rembrandt.''}
    \label{fig: amb a}
\end{figure*}

%% file: figure/amb_b.tex
\begin{figure*}
    \centering
    \includegraphics[width=\linewidth]{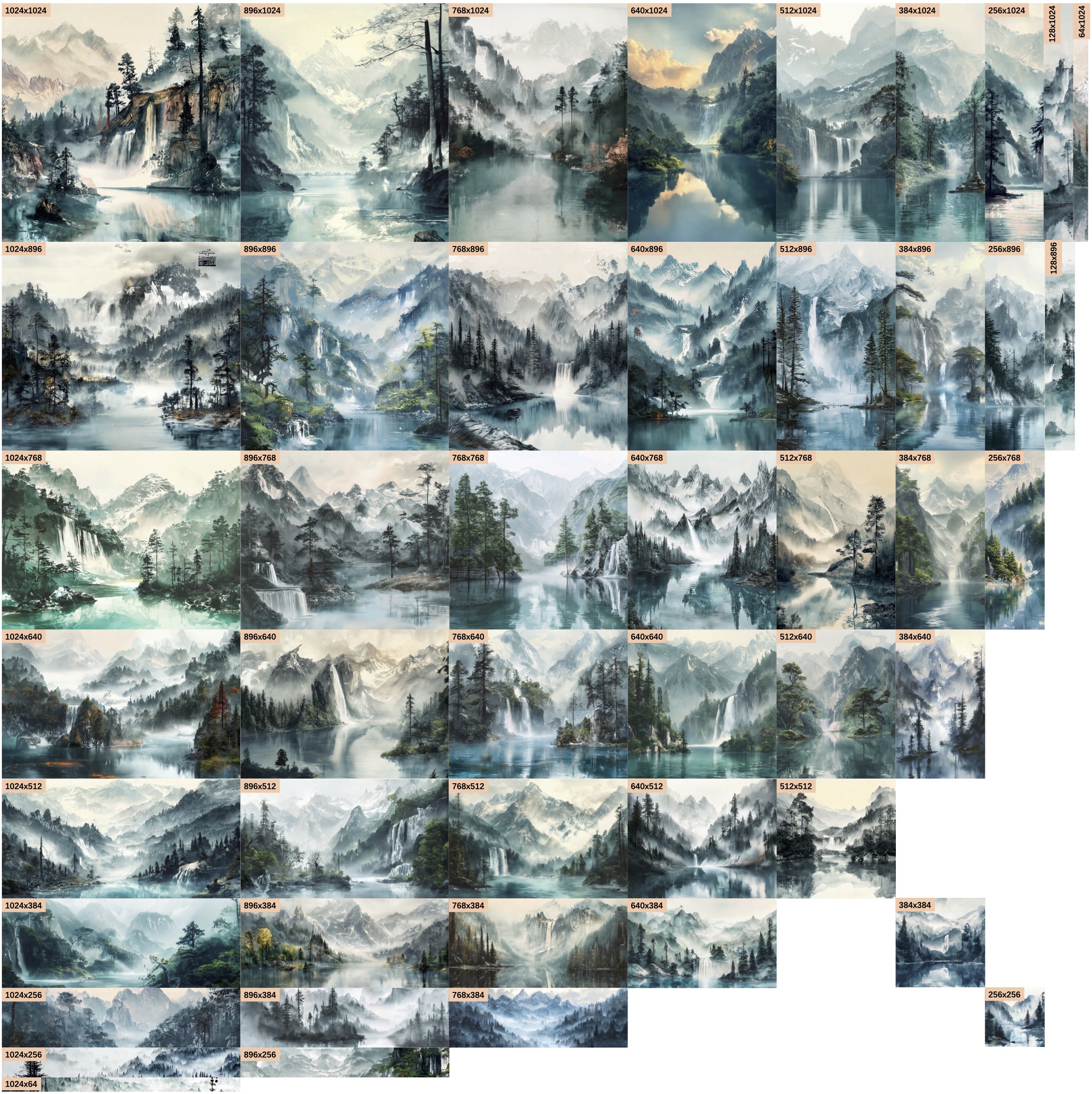}
    \caption{D-JEPA$\cdot$T2I can generate arbitrary aspect ratios and continuous resolutions with VoPE. Prompt: ``A serene mountain landscape in the style of a Chinese ink painting, with a waterfall cascading down into a crystal-clear lake surrounded by ancient pines.''}
    \label{fig: amb b}
\end{figure*}

%% file: figure/amb_c.tex
\begin{figure*}
    \centering
    \includegraphics[width=\linewidth]{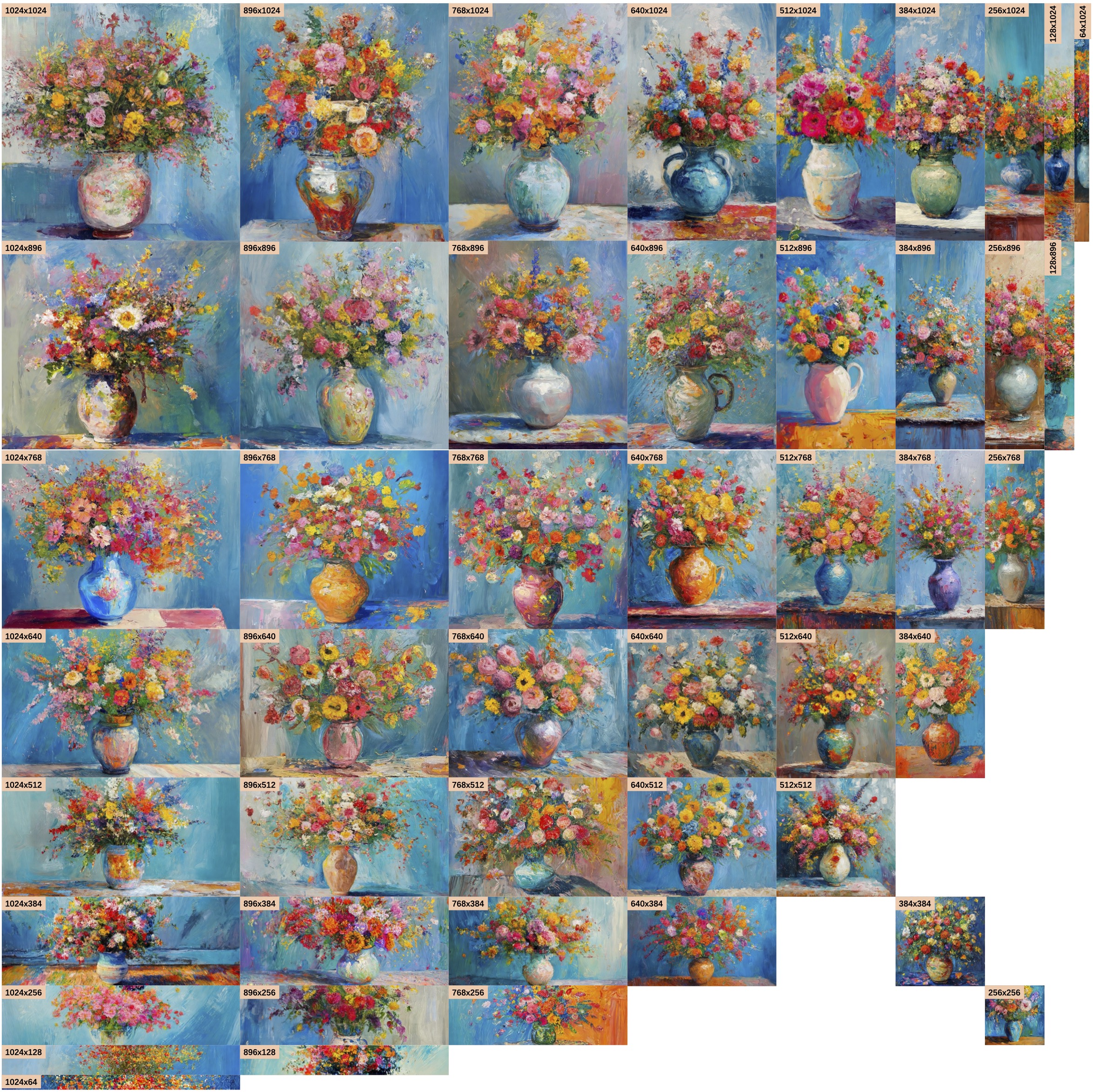}
    \caption{D-JEPA$\cdot$T2I can generate arbitrary aspect ratios and continuous resolutions with VoPE. Prompt: ``A still life of a vase overflowing with vibrant flowers, painted in bold colors and textured brushstrokes, reminiscent of van Gogh's iconic style.''}
    \label{fig: amb c}
\end{figure*}

%% file: figure/amb_d.tex
\begin{figure*}
    \centering
    \includegraphics[width=\linewidth]{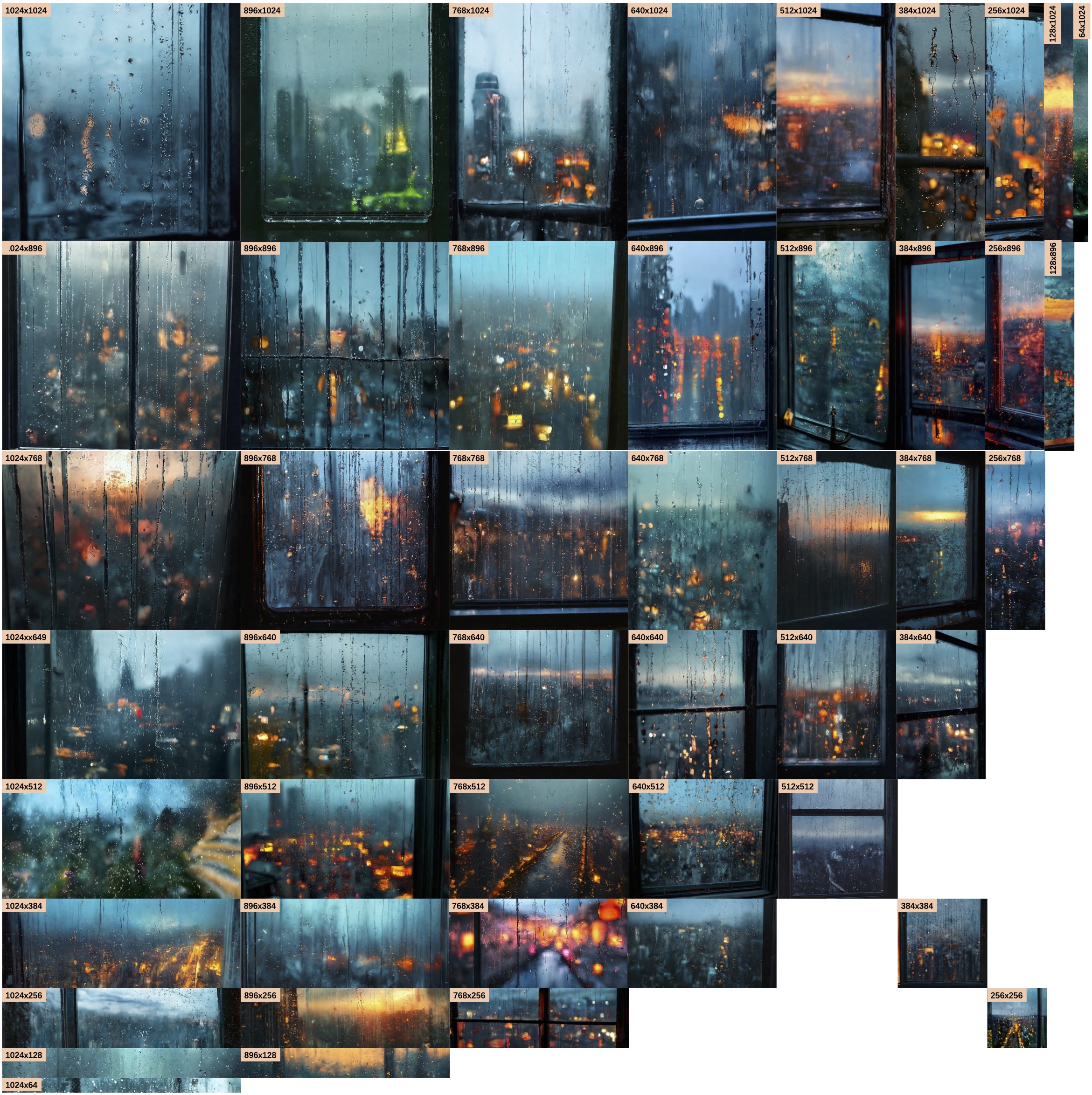}
    \caption{D-JEPA$\cdot$T2I can generate arbitrary aspect ratios and continuous resolutions with VoPE. Prompt: ``A window with raindrops trickling down, overlooking a blurry city.''}
    \label{fig: amb d}
\end{figure*}

%% file: figure/bias.tex
\begin{figure}
    \centering
    \includegraphics[width=\linewidth]{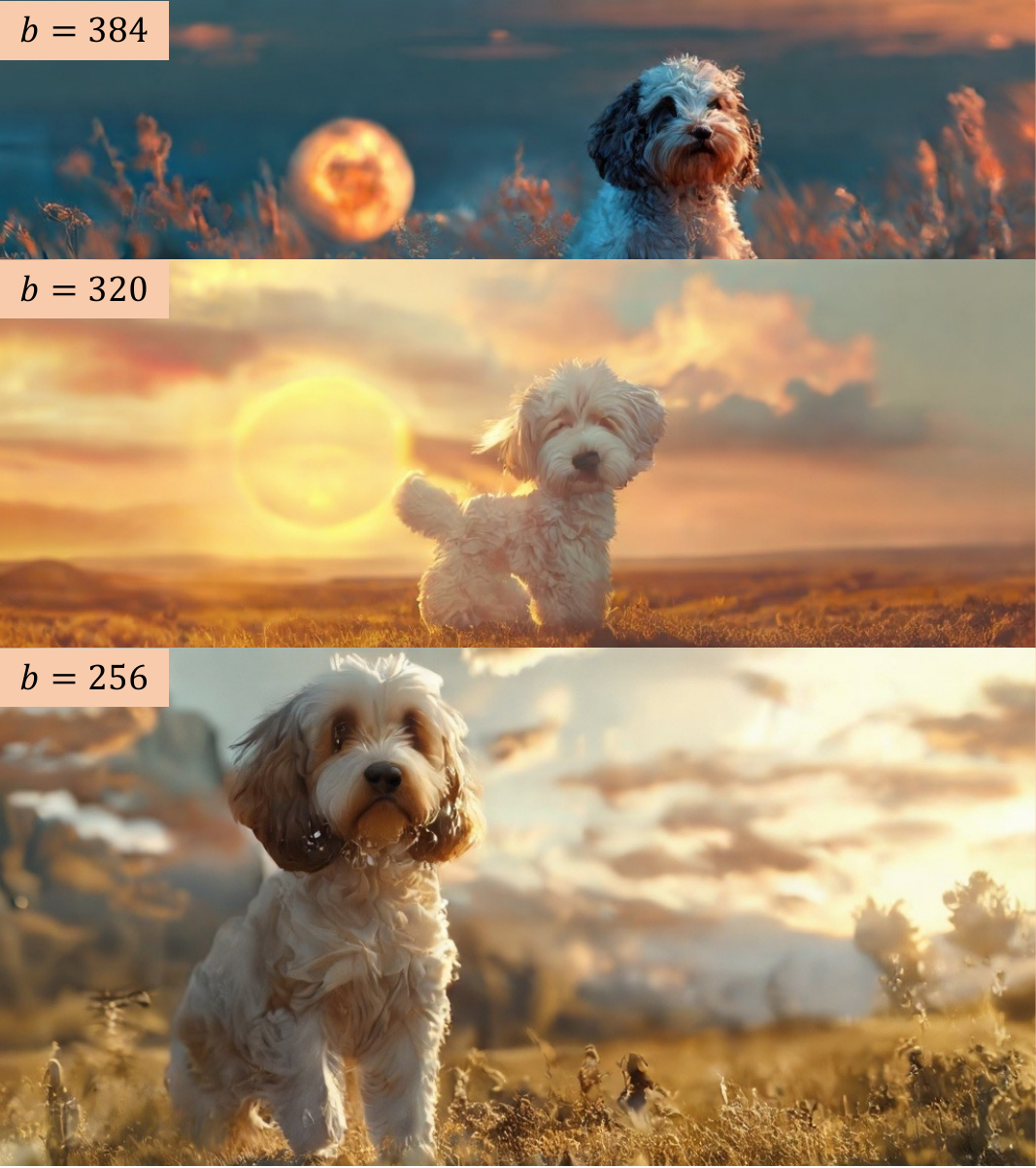}
    \caption{Layout control by relative positional offset $b$. By adjusting $b$, we can generate more desirable layouts for selection.}
    \label{fig: bias}
\end{figure}

%% file: sec/sup/comparison.tex
\section{Visual Comparison with Other Methods}

In this section, we present a visual comparison of the images generated by D-JEPA$\cdot$T2I with those produced by other state-of-the-art methods. This allows for a qualitative assessment of the capabilities and advantages of D-JEPA$\cdot$T2I in generating high-quality images across various scenarios and styles. Through side-by-side comparisons, we can evaluate how well each method captures details, maintains image consistency, and handles different resolutions and aspect ratios.
\input{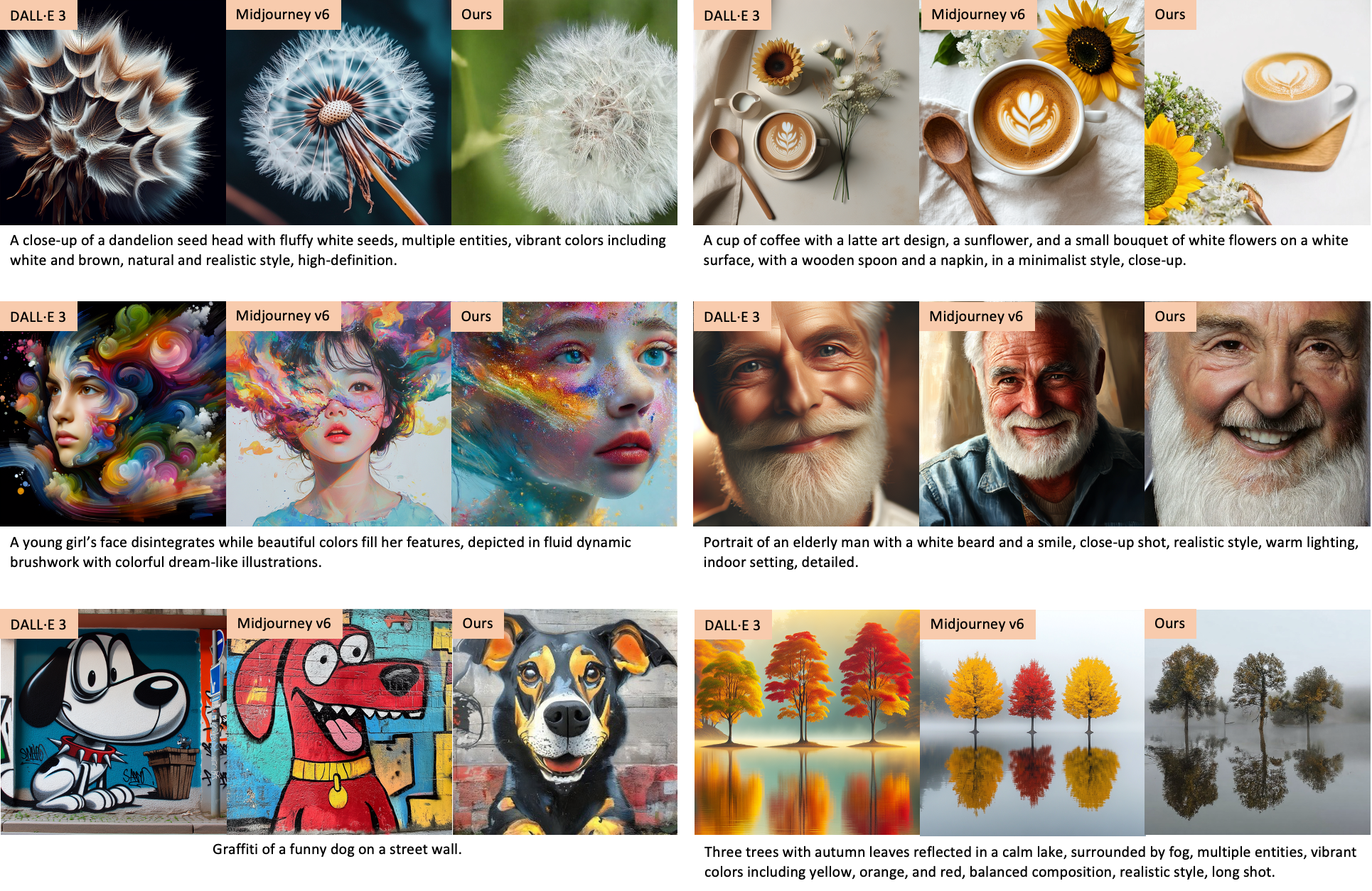}
\input{figure/cmp_fluid}
\input{figure/cmp_lumina}
\input{figure/cmp_pixelart}
\input{figure/cmp_hunyuandit}
\input{figure/cmp_transfusion}

In Fig.~\ref{fig: cmp ots}, ~\ref{fig: cmp fluid}, ~\ref{fig: cmp lumina}, ~\ref{fig: cmp pixelart}, ~\ref{fig: cmp hunyuandit} and ~\ref{fig: cmp transfusion}, the examples displayed highlight specific attributes of image quality, such as clarity, color accuracy, and complexity of details. These comparisons help demonstrate the strengths and potential limitations of different models in addressing diverse visual tasks.

%% file: figure/cmp_ots.tex
\begin{figure*}
    \centering
    \includegraphics[width=\linewidth]{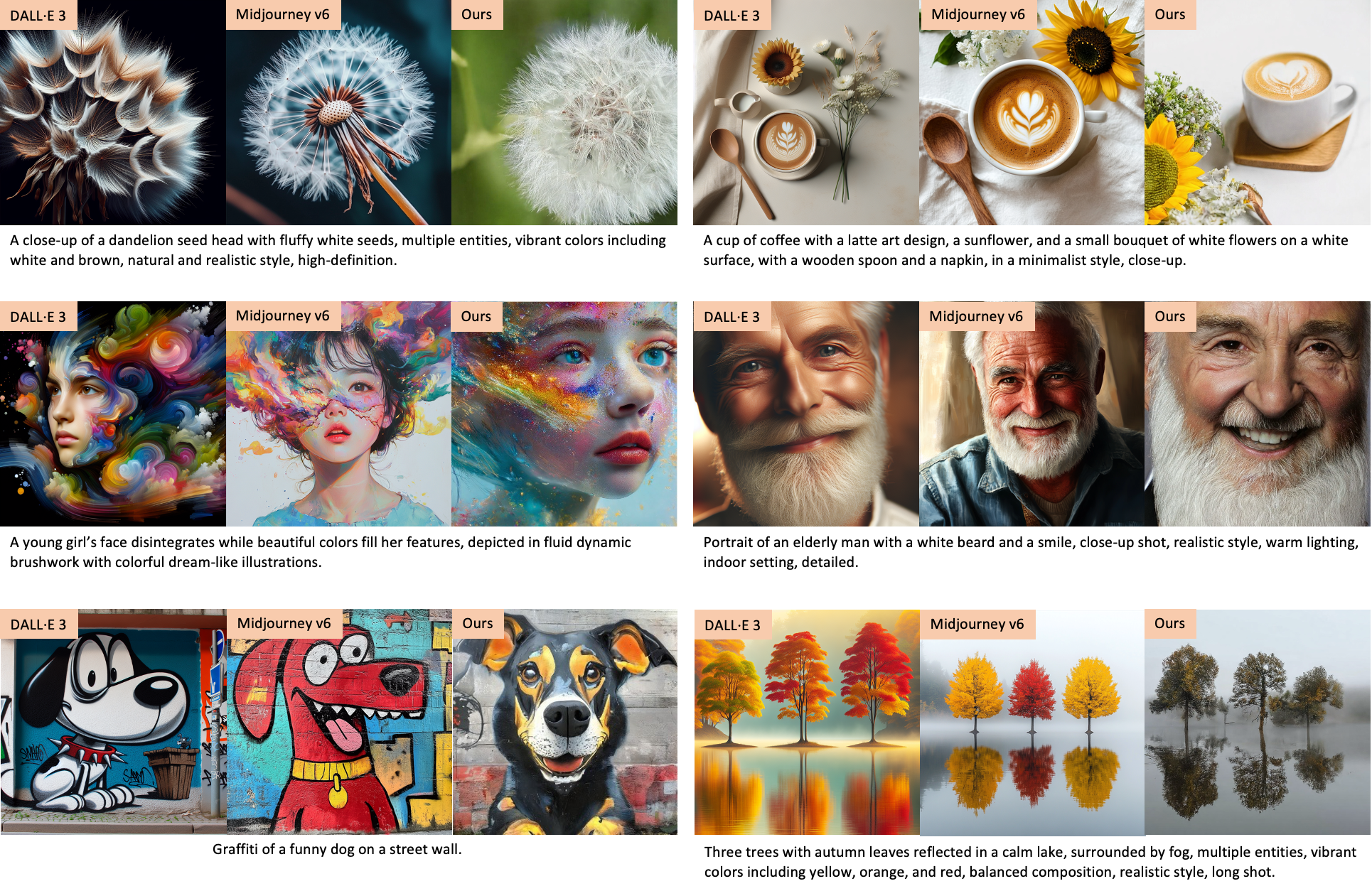}
    \caption{Visual comparison among commercial models, \textit{i.e.,} DALL$\cdot$E 3~\citep{betker2023improving}, Midjourney v6~\citep{midjourneyv6}, and D-JEPA$\cdot$T2I.}
    \label{fig: cmp ots}
\end{figure*}

%% file: figure/cmp_fluid.tex
\begin{figure*}
    \centering
    \includegraphics[width=\linewidth]{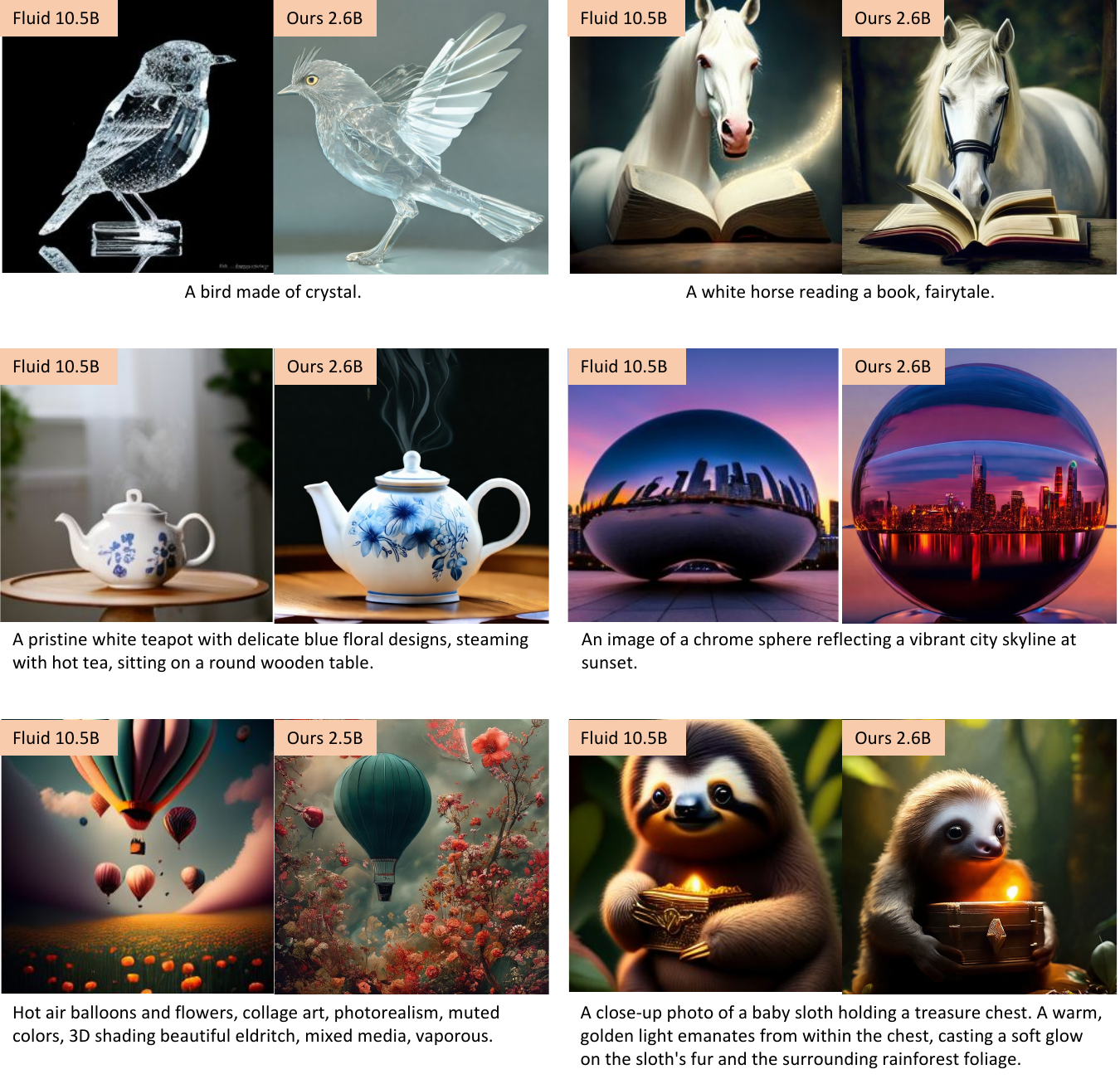}
    \caption{Visual comparison between Fluid~\citep{fan2024fluid} and D-JEPA$\cdot$T2I.}
    \label{fig: cmp fluid}
\end{figure*}

%% file: figure/cmp_lumina.tex
\begin{figure*}
    \centering
    \includegraphics[width=\linewidth]{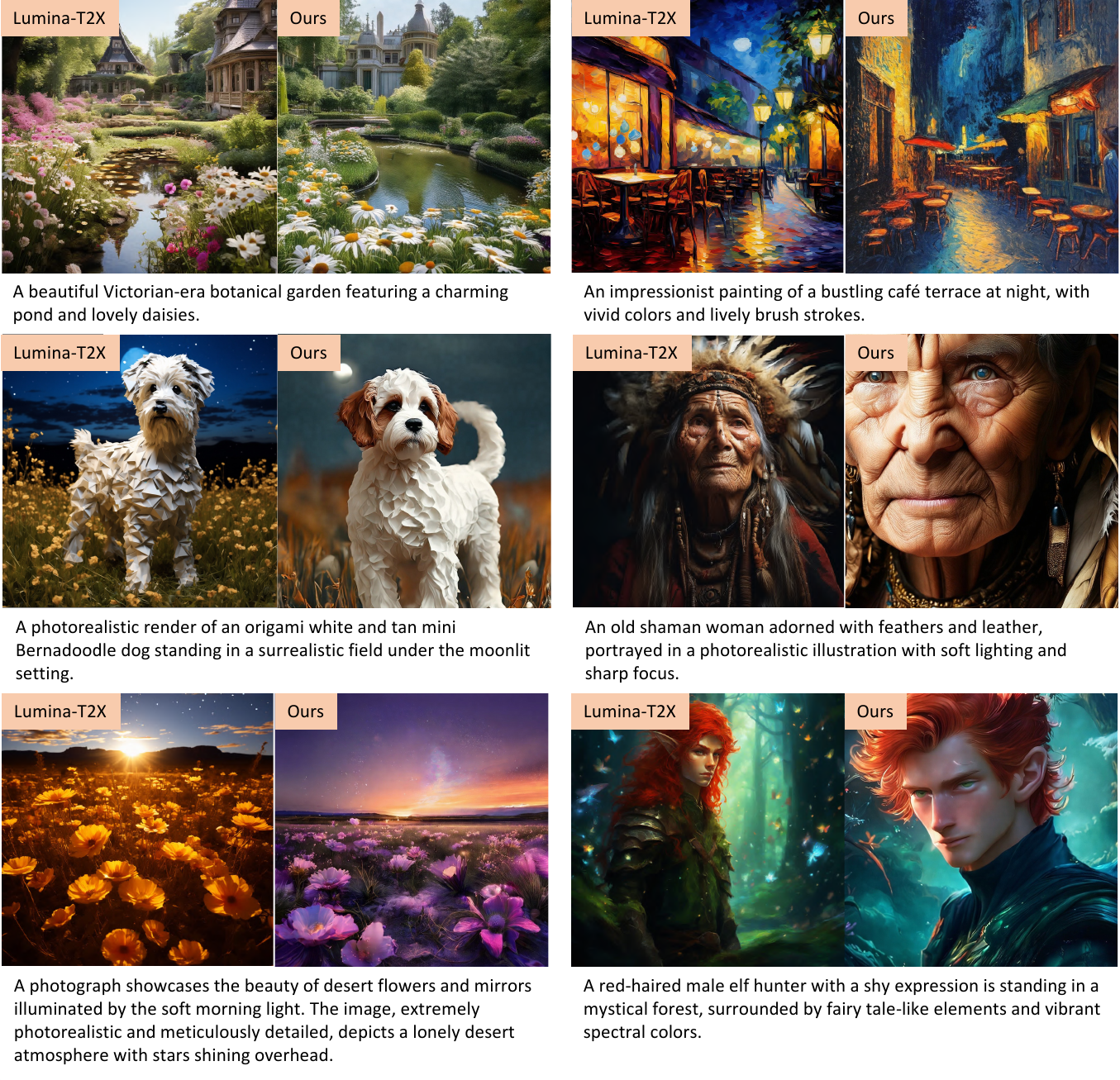}
    \caption{Visual comparison between Lumina-T2X~\citep{gao2024lumina} and D-JEPA$\cdot$T2I.}
    \label{fig: cmp lumina}
\end{figure*}

%% file: figure/cmp_pixelart.tex
\begin{figure*}
    \centering
    \includegraphics[width=\linewidth]{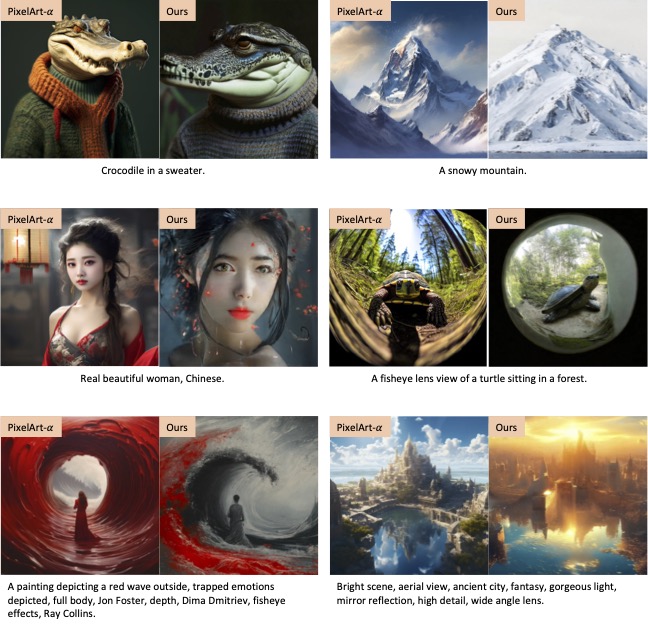}
    \caption{Visual comparison between PixelArt-$\alpha$~\citep{chen2023pixart} and D-JEPA$\cdot$T2I.}
    \label{fig: cmp pixelart}
\end{figure*}

%% file: figure/cmp_hunyuandit.tex
\begin{figure*}
    \centering
    \includegraphics[width=\linewidth]{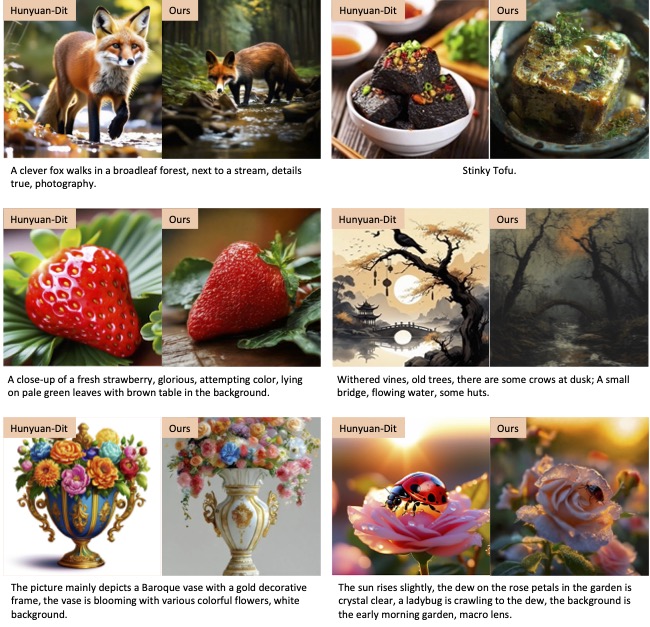}
    \caption{Visual comparison between HunyunDit~\citep{li2024hunyuan} and D-JEPA$\cdot$T2I.}
    \label{fig: cmp hunyuandit}
\end{figure*}

%% file: figure/cmp_transfusion.tex
\begin{figure*}
    \centering
    \includegraphics[width=\linewidth]{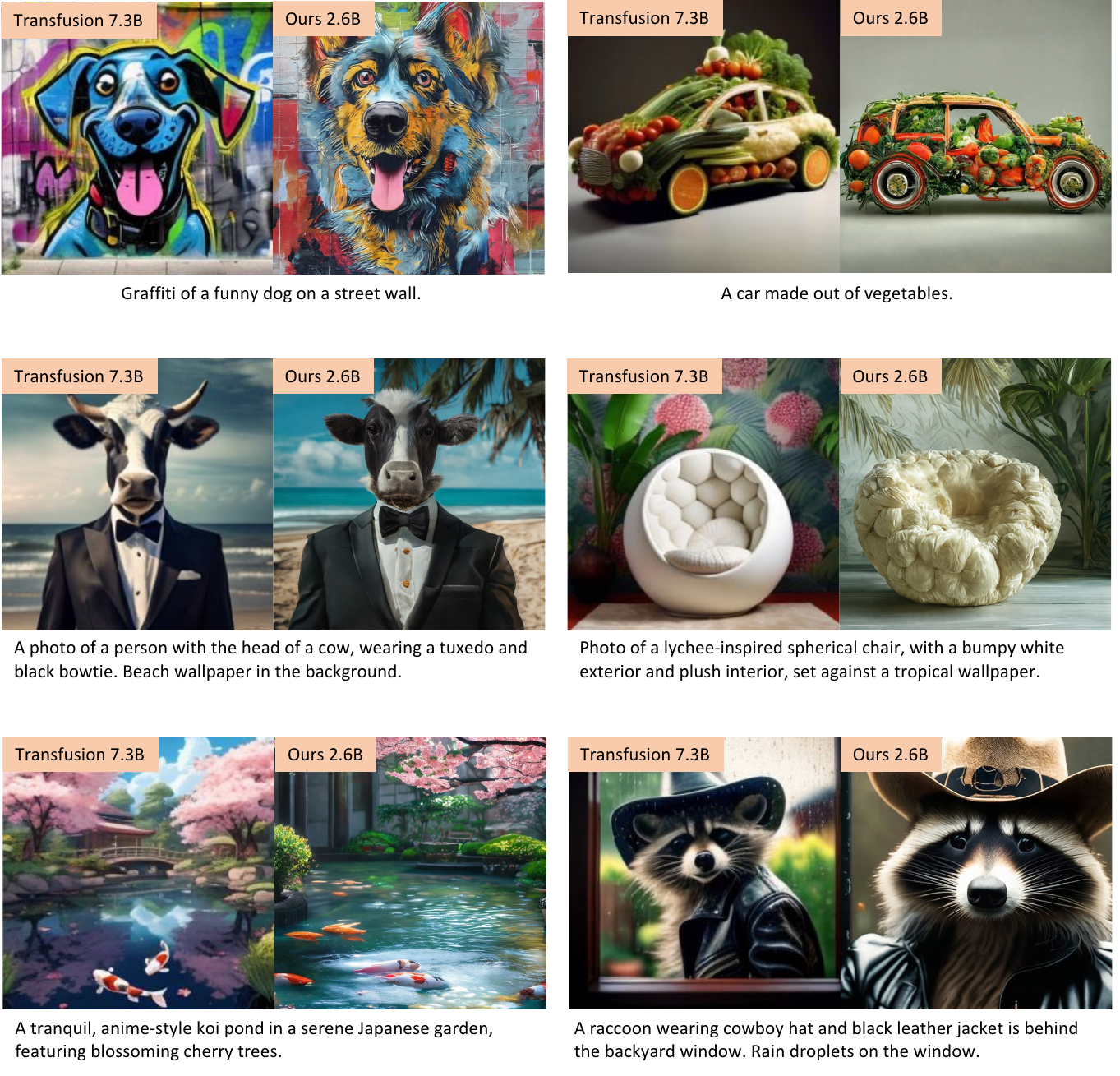}
    \caption{Visual comparison between Transfusion~\citep{zhou2024transfusion} and D-JEPA$\cdot$T2I.}
    \label{fig: cmp transfusion}
\end{figure*}

%% file: sec/sup/limitation.tex
\section{Limitation}

\paragraph{Limitations due to training data.} In our training process, we excluded images containing text, which limits D-JEPA$\cdot$T2I's capability to perform text generation tasks. This may hinder its application in contexts where text is critical, such as in advertisements and posters. Moreover, D-JEPA$\cdot$T2I currently only supports English prompts and cannot yet accommodate prompts in other languages like Chinese. Its performance is also less than optimal when generating high-resolution images (4k). However, these limitations can be addressed by integrating more diverse data types and improving data quality.

\paragraph{Limitations due to inference efficiency.} While D-JEPA$\cdot$T2I utilizes a next set-of-tokens prediction strategy for training and inference, it relies on bi-directional attention. This requirement precludes employing a key-value cache mechanism, which can speed up inference as seen with causal attention. Generating a 1k resolution image typically takes around a minute, imposing constraints on its use in highly interactive applications. Enhancements in inference efficiency could be achieved through feature compression techniques, as discussed in ~\citet{chen2024pixart}.

\paragraph{Limitations due to model size.} Our current model is trained with 2.6B parameters, and we have not yet investigated performance at larger scales, such as over 10B parameters. Although the 2.6B model delivers satisfactory image generation, it still faces challenges in accurately generating specific entities and nouns, especially for people and places. Its performance is also lacking for prompts involving numerical, spatial relationships, and complex scenarios. Increasing the model size is expected to significantly improve its overall performance, as evidenced by findings in ~\citet{esser2024scaling}.

%% file: sec/sup/prompt.tex
\section{Prompts}
All prompts are listed in the order of the corresponding images in the picture from top to bottom, from left to right.

\paragraph{The prompts used in Fig.~\ref{fig: teaser}.}
\begin{itemize}
    \item A young girl's face disintegrates while beautiful colors fill her features, depicted in fluid dynamic brushwork with colorful dream-like illustrations.
    \item Close-up portrait of a young woman with long brown hair, neutral expression, soft lighting, minimalistic style, indoor setting, eye level.
    \item Stars, water, brilliantly, gorgeous large scale scene, a little girl, in the style of dreamy realism.
    \item Portrait of an elderly man with a white beard and a smile, close-up shot, realistic style, warm lighting, indoor setting, detailed.
    \item A smiling elderly man with gray hair and beard, wearing a black and white checkered shirt, close-up shot, neutral background, realistic style, eye level.
    \item Art collection style and fashion shoot, in the style of made of glass, dark blue and light pink, paul rand, solarpunk, camille vivier, beth didonato hair, barbiecore, hyper-realistic.
    \item A watercolor portrait of a Terrier dog, smiling and making a cute facial expression while looking at the camera, in Pixar style.
    \item Nature vs. human nature, surreal, UHD, 8k, hyper details, rich colors, photograph.
    \item Graffiti of a funny dog on a street wall.
    \item Detailed pen and ink drawing of a massive complex alien space ship above a farm in the middle of nowhere.
    \item A close-up of a chibi-style figurine with a brown beard, blue eyes, and a green shirt with a yellow emblem, holding a gray stuffed animal, standing on a white surface with a blurred background, realistic style, warm lighting, detailed.
    \item A pair of old boots covered in mud.
    \item An entire universe inside a bottle sitting on the shelf at walmart on sale.
    \item Film still of a long-legged cute big-eye anthropomorphic cheeseburger wearing sneakers relaxing on the couch in a sparsely decorated living room.
    \item Human life depicted entirely out of fractals.
    \item Three trees with autumn leaves reflected in a calm lake, surrounded by fog, multiple entities, vibrant colors including yellow, orange, and red, balanced composition, realistic style, long shot.
    \item Grilled skewers of chicken and bell peppers, multiple entities, vibrant colors including yellow, red, and green, close-up, realistic style, outdoor setting, daylight.
    \item A panoramic view of a sunflower field with vibrant yellow and green hues, rolling hills in the background, a small village with red-roofed buildings, and a clear blue sky, captured in a realistic style with a wide shot perspective.
    \item Two small boats docked at a wooden pier on a calm lake, with a lush green forest in the background, warm lighting, realistic style, medium shot.
    \item Three spheres made of glass falling into ocean. Water is splashing. Sun is setting.
    \item A pristine white teapot with delicate blue floral designs, steaming with hot tea, sitting on a round wooden table.
    \item A chocolate cake with a dusting of powdered sugar, cut into eight slices, placed on a piece of parchment paper on a wooden surface, with a glass of honey and a fork on the side, realistic style, close-up, warm lighting.
    \item Stinky Tofu.
    \item Grilled skewers of chicken and bell peppers, multiple entities, vibrant colors including yellow, red, and green, close-up, realistic style, outdoor setting, daylight.
\end{itemize}

\paragraph{The prompts used in Fig.~\ref{fig: failure}.}
\begin{itemize}
    \item A close-up of a metal staircase with a complex, interlocking design, featuring multiple levels and railings, in a monochromatic color scheme, with a focus on the geometric patterns and structural details, in a photorealistic style.
    \item A geometric metal structure with a series of intersecting lines and angles, creating a dynamic and abstract pattern, viewed from a low angle, minimalist style, black and white photograph.
    \item A collection of fabric swatches in various colors and patterns, including shades of brown, green, gold, and gray, arranged in a grid pattern on a white background, with a focus on texture and detail, in a realistic style.
    \item A desert scene with a sand dune leading to a bright light source at the horizon, the Milky Way visible in the night sky, long exposure, minimalistic style, black and white.
    \item A white ceramic bowl with a textured pattern, placed on a woven bamboo mat, on a dark wooden table, minimalistic style, close-up.
    \item A white ceramic bowl with a textured pattern, placed on a woven bamboo mat, on a dark wooden table, minimalistic style, close-up.
    \item A close-up of a single water droplet in mid-air, with a splash of water droplets surrounding it, against a white background, realistic style, high-definition.
    \item A geometric abstract composition featuring a mix of black and white polygons, with a focus on the spatial relationship between the shapes, minimalistic style, and a modern, clean aesthetic.
\end{itemize}

\paragraph{The prompts used in Fig.~\ref{fig: 4k}.}
\begin{itemize}
    \item A black and white photograph of a woman wearing a wide-brimmed hat, with a close-up shot of her face revealing delicate facial features, high contrast.
    \item Close-up of a brown horse with a white blaze on its face, standing in a misty field, realistic style, detailed, natural lighting, outdoor setting, long shot.
    \item Single tree with vibrant yellow leaves standing in a calm lake, surrounded by mountains, reflected in the still water, autumnal colors, long shot, photorealistic style, natural lighting.
\end{itemize}